\documentclass[sn-nature]{sn-jnl}

\usepackage{graphicx}%
\usepackage{multirow}%
\usepackage{amsmath,amssymb,amsfonts}%
\usepackage{amsthm}%
\usepackage{mathrsfs}%
\usepackage[title]{appendix}%
\usepackage{xcolor}%
\usepackage{textcomp}%
\usepackage{manyfoot}%
\usepackage{booktabs}%
\usepackage{algorithm}%
\usepackage{algorithmicx}%
\usepackage{algpseudocode}%
\usepackage{listings}%

\usepackage{natbib}
\usepackage{subcaption}

\theoremstyle{thmstyleone}%
\newtheorem{theorem}{Theorem}
\newtheorem{proposition}[theorem]{Proposition}%

\theoremstyle{thmstyletwo}%
\newtheorem{example}{Example}%
\newtheorem{remark}{Remark}%

\theoremstyle{thmstylethree}%
\newtheorem{definition}{Definition}%

\begin{document}

\title[Article Title]{Enhancing Unimodal Latent Representations in Multimodal VAEs through Iterative Amortized Inference}


\author*[1]{\fnm{Yuta} \sur{Oshima}}\email{yuta.oshima@weblab.t.u-tokyo.ac.jp}

\author[1]{\fnm{Masahiro} \sur{Suzuki}}

\author[1]{\fnm{Yutaka} \sur{Matsuo}}

\affil*[1]{\orgdiv{Graduate School of Engineering}, \orgname{, The University of Tokyo}, \orgaddress{\street{7-3-1 Hongo}, \city{Bunkyo-ku}, \postcode{113-8656}, \state{Tokyo}, \country{Japan}}}




\abstract{
In recent years, deep generative models for multimodal data have gained significant attention. 
Among these, multimodal variational autoencoders (VAEs) have emerged as a promising approach, aiming to capture a shared latent representation by integrating information across different modalities through their inference models. 
A primary challenge for multimodal VAEs is accurately inferring representations from arbitrary subsets of modalities after learning a multimodal inference model. 
Naively, this would require training \( 2^M \) different inference networks (\(M\) is \# of modalities) to handle every possible combination of modalities, which is infeasible for a large number of modalities.
Mixture-based models address this challenge by requiring only as many inference models as there are modalities, aggregating unimodal inferences to perform multimodal inference. 
However, when modalities are missing, these models suffer from information loss, particularly of modality-specific information, leading to deteriorated inference performance. 
Alternatively, alignment-based multimodal VAEs aim to align unimodal inference models with a multimodal inference model by minimizing the Kullback–Leibler (KL) divergence between them. 
Yet, the multimodal amortized inference, which is alignment source in these models inherently suffers from amortization gaps, preventing it from perfectly approximating the true inference and compromising the accuracy of unimodal inference.
To address both issues, we introduce an iterative amortized inference mechanism within the multimodal VAE framework, termed multimodal iterative amortized inference. 
By iteratively refining the multimodal inference using all modalities, this method overcomes the information loss due to missing modalities in mixture-based models and minimizes the amortization gap in alignment-based models. 
Furthermore, by aligning the unimodal inference to approximate this refined multimodal posterior, we obtain unimodal inferences that effectively incorporate multimodal information while requiring only unimodal inputs at inference time.
Experimental results on two benchmark datasets demonstrate that the proposed method improves the performance of the inference itself, suggested by higher linear classification accuracy and cosine similarity, and that the learned representations effectively capture the distributions of other modalities, as indicated by lower Fréchet Inception Distance (FID) scores in cross-modal generation.
This indicates that the proposed approach significantly enhances the inferred representations from unimodal inputs.}




\maketitle

\section{Introduction}\label{sec1}

Humans acquire multimodal information from the world, deepening their understanding. This highlights the importance of multimodal data processing, a crucial aspect of artificial intelligence aimed at comprehending the complexities of our environment~\citep{baltrusaitis18}.
In self-supervised multimodal learning, variational autoencoders (VAEs)~\citep{kingma13} have gained prominence, leading to the development of their multimodal variants, termed multimodal VAEs~\citep{suzuki16, suzuki22}. 
VAEs are adept at encoding inputs into latent representations by learning the inference of latent variables through their encoder-decoder architecture. When provided with all modalities, they can infer latent representations from multimodal inputs.

However, a key challenge in multimodal VAEs arises when a modality is missing; the inference collapses if we attempt to infer the latent representation from only a subset of modalities~\citep{suzuki16}. 
This collapse occurs because the inference network is approximated by a neural network that expects all modalities as input. 
To address this issue, one might consider designing separate inference networks for each possible subset of modalities to handle missing inputs. 
However, to infer from any combination of $M$ modalities, a naive approach would require training $2^M$ encoders, leading to an exponential increase in training cost with the number of modalities.

Recent models, known as mixture-based models, such as MVAE~\citep{wu18}, MMVAE~\citep{shi19}, and MoPoE-VAE~\citep{sutter21}, allow for inference from any combination of modalities using only $M$ inference models by aggregating unimodal inferences through mechanisms such as PoE~\citep{hinton02}, MoE, and MoPoE (Figure~\ref{fig:mixture-based}). 
However, these models rely on subsampling modalities during training, which imposes a theoretical limitation on the performance of inference from subsets and leads to information loss due to missing modalities. Experimental results have demonstrated that cross-modal generation performance from these inferred representations deteriorates~\citep{daunhawer21}.

Some studies called alignment-based multimodal VAEs address this by learning inferences from all modalities (i.e., multimodal inference) and optimizing unimodal inferences to minimize the Kullback–Leibler (KL) divergence between them and the multimodal inference~\citep{suzuki16, hwang21} (Figure~\ref{fig:alignment-based}). 
This approach avoids the theoretical limitations associated with subsampling in mixture-based models. 
Alignment-based multimodal VAEs utilize a two-step approximation for unimodal inferences: (1) approximating the true multimodal inference with a neural network through amortized variational inference, and (2) approximating the multimodal inference using unimodal inferences. 
Ideally, unimodal inferences should closely approximate the true multimodal inference. 
However, the quality of this approximation depends heavily on the effectiveness of the multimodal amortized inference (the first-step approximation). 
It is known that amortized inference, typically using neural network approximations, inherently has a gap between it and the true inference, referred to as the amortization gap~\citep{cremer18}. Consequently, the performance of unimodal inference is also suboptimal.

To address these issues, we propose utilizing iterative amortized inference~\citep{cremer18}, a method originally introduced to reduce the amortization gap in amortized inference.  
Multimodal iterative amortized inference iteratively improves the unimodal inference by ascending the gradient of the multimodal evidence lower bound (ELBO), which requires information from all modalities for its calculation (see the upper part of Figure~\ref{iai_fig}). 
By improving the unimodal inference using this method via the multimodal ELBO, we can recover information from other modalities while filling the amortization gap. 
In doing so, our method overcomes both the information loss due to missing modalities in mixture-based models and the amortization gap present in alignment-based models.

However, multimodal iterative amortized inference alone cannot perform inference from truly unimodal inputs. 
This is because the calculation of the multimodal ELBO requires information from all modalities. 
To bridge this gap, we follow the approach of alignment-based models by minimizing the Kullback–Leibler (KL) divergence between the unimodal inference and the multimodal iterative amortized posterior (see the lower part of Figure~\ref{iai_fig}). 
By aligning the unimodal inference to approximate the multimodal iterative amortized posterior, we can obtain unimodal inferences that are sourced from an inference process that overcomes both the information loss due to missing modalities and the amortization gap.

We conducted experiments on standard benchmarks—the MNIST-SVHN-Text dataset~\citep{sutter20} and the Caltech Birds (CUB) dataset~\citep{wah2011caltech}. 
Our results demonstrated that the proposed method significantly improves the performance of the inference itself, as evidenced by higher linear classification accuracy and greater cosine similarity of latent representations. 
Additionally, the representations learned by our method effectively capture the distributions of other modalities, which is reflected in lower Fréchet Inception Distance (FID) scores~\citep{heusel17} in cross-modal generation. 
This indicates that our method not only minimizes the amortization gap but also overcomes the information loss due to missing modalities observed in mixture-based models. 

\section{Preliminary}

\subsection{Multimodal VAEs}

The purpose of multimodal VAEs is to maximize the likelihood $p_{\theta}(X)$ with respect to the parameter $\theta$ under the given $M$ types of multimodal inputs $X=\{\mathbf{x}_{m}\}_{m=0}^{M-1}$, where the marginal log-likelihood is defined as $p_{\theta}(X)=\int p_{\theta}(X|\mathbf{z}) d\mathbf{z} = \int\prod_{m:\mathbf{x}_m\in X} p_{\theta}(\mathbf{x}_m|\mathbf{z}) d\mathbf{z}$~\citep{suzuki22}. Here, $\mathbf{z}$ is the latent variable, the shared representation in multimodal VAE. In practice, since the log-likelihood cannot be directly evaluated, the evidence lower bound (ELBO) of the log-likelihood is used as the objective function for learning:
\begin{equation}
\mathcal{L}(\theta, \phi; X) = \mathbb{E}_{q_{\phi}(\mathbf{z}|X)}[\log p_{\theta}(X|\mathbf{z})] - D_{\text{KL}}[q_{\phi}(\mathbf{z}|X)||p(\mathbf{z})],
\label{eq:elbo}
\end{equation}
where $q_{\phi}(\mathbf{z}|X)$ is the inference model and is an approximation of the true posterior distribution $p_{\theta}(\mathbf{z}|X)$. Considering VAEs as autoencoders, the inference model is also called an encoder. A major challenge in multimodal VAEs is that the above maximization only learns the inference from all modalities, so it cannot perform the inference from an arbitrary modality $\mathbf{x}_m$.
Therefore, our objective is to obtain a unimodal inference $q_{\phi_m}(\mathbf{z}|\mathbf{x}_m)$ that approximates the true multimodal inference $p_{\theta}(\mathbf{z}|X)$.

\subsubsection*{Mixture-based multimodal VAEs}
In mixture-based multimodal VAEs, it is possible to learn inference from any modality using only $M$ encoders (Figure~\ref{fig:mixture-based}). Naively, $2^M$ encoders would be required to perform inference from any modality, and as the number of modalities increases, the computational cost of learning becomes excessive. In mixture-based methods such as MVAE~\citep{wu18}, MMVAE~\citep{shi19}, MoPoE-VAE~\citep{sutter21}, the number of inferences can be limited to the number of modalities by aggregating the encoders of the modalities to be used for inference through techniques such as PoE, MoE, or MoPoE, consequently reducing the required number of encoders.

As an aggregation method, MVAE uses the product of experts (PoE)~\citep{hinton02}:
\begin{equation}
q_{\phi}^{\text{PoE}}(\mathbf{z}|X_S) \equiv p(\mathbf{z}) \prod_{m:\mathbf{x}_m \in X_S} q_{\phi_m}(\mathbf{z}|\mathbf{x}_m),
\end{equation}
and MVAE uses the mixture of experts (MoE):
\begin{equation}
q_{\phi}^{\text{MoE}}(\mathbf{z}|X_S) \equiv \sum_{m:\mathbf{x}_m \in X_S} q_{\phi_m}(\mathbf{z}|\mathbf{x}_m),
\end{equation}
where $S$ represents a subset of $\{1,..,M\}$.
The mixture of the product of experts (MoPoE) is a generalized method that combines both PoE and MoE and can be written as:
\begin{equation}
    \begin{gathered}
        q_{\phi}^{\text{MoPoE}}(\mathbf{z}|X_S) \equiv \sum_{S:X_S \in \mathcal{P}(X)} \omega_S \left( \prod_{m:\mathbf{x}_m \in X_S} p(\mathbf{z}) q_{\phi_m}(\mathbf{z}|\mathbf{x}_m) \right) \\
        = \sum_{S:X_S \in \mathcal{P}(X)} \omega_S q_{\phi}^{\text{PoE}}(\mathbf{z}|X_S),
    \end{gathered}
\end{equation}
where $\sum_{S:X_S \in \mathcal{P}(X)} \omega_S = 1$, $\omega_S \in [0, 1]$, and where $\mathcal{P}(X)$ is the power set of $X$.
Using the posterior distribution obtained by MoPoE, the lower bound for Equation~\eqref{eq:elbo} becomes:
\begin{equation}
    \begin{gathered}
        \mathcal{L}(\theta, \phi; X) \geq\sum_{S:X_S \in P(X)} \omega_S \left( \mathbb{E}_{q_{\phi}^{\text{PoE}} (\mathbf{z}|X_S)}[\log p_{\theta}(X|\mathbf{z})] \right.  \left. - D_{\text{KL}}[q_{\phi}^{\text{PoE}} (\mathbf{z}|X_S)||p(\mathbf{z})] \right) \\
        = \mathcal{L}_M(\theta, \phi; X).
    \end{gathered}
\end{equation}

However, it is known that the lower bound in MoPoE, or the multimodal ELBO, $\mathcal{L}_{M}(\theta, \phi; X)$ is constrained by the following inequality with the expected value of the marginal log-likelihood~\citep{daunhawer21}:
\begin{equation}
\mathbb{E}_{p_d(X)}[\log p_{\theta}(X)] \geq \Delta(X) + \mathbb{E}_{p_d(X)}[\mathcal{L}_{M}(\theta, \phi; X)],
\end{equation}
where $p_d(X)$ is data distribution and where
\begin{equation}
\Delta(X) \equiv \sum_{S:X_S \in P(X)} w_S H(X_{\{1,\ldots,M\}\setminus S}) \,|\, X_S).
\end{equation}

Since $\Delta(X)$ does not depend on the learning parameters, it means that no matter how much the lower bound in MoPoE is maximized in terms of the learning parameters, it cannot approach the expected value of the marginal log-likelihood by the amount of $\Delta(X)$. Here, $\Delta(X)$ represents the size of the information specific to $X_{\{1,\ldots,M\}\setminus S}$ (which $X_S$ does not possess). This problem arises from the subsampling of the modality in the calculation of multimodal ELBO, and if the difference in information between $X_S$ and $X_{\{1,\ldots,M\}\setminus S}$ is significant, the inference accuracy of the posterior distribution by MoE or MoPoE will be reduced. In practice, even with two modalities, it has been confirmed that MoE and MoPoE can fail to infer from a single modality or perform cross-modal generation.

\subsubsection*{Alignment-based multimodal VAEs}
Alignment-based multimodal VAEs are types of multimodal VAEs that enables the acquisition of shared representations from a single modality~\citep{suzuki16, hwang21}. 
There models achieve this by preparing an inference model $q_{\lambda_m}(\mathbf{z}|\mathbf{x}_m)$ that takes only one modality as input, in addition to the inference $q_{\phi}(\mathbf{z}|X)$ from all modalities, and aligning them (Figure~\ref{fig:alignment-based}). 
Although alignment-based multimodal VAEs were not explicitly proposed to overcome the theoretical limitations of mixture-based multimodal VAEs, it avoids modality subsampling in the calculation of the ELBO, thus circumventing the theoretical constraints. 
The objective function of alignment-based multimodal VAEs is described below:
\begin{equation}
\label{eq:mvtcae}
\mathbb{E}_{q_{\phi}(\mathbf{z}|X)}[\log p_{\theta}(X|\mathbf{z})] 
-D_{\text{KL}}[q_{\phi}(\mathbf{z}|X)||p(\mathbf{z})] 
-\sum_{m=1}^{M} \pi_m D_{\text{KL}}(q_{\phi}(\mathbf{z}|X)||q_{\lambda_m}(\mathbf{z}|\mathbf{x}_m)), 
\end{equation}
where $\sum_m \pi_m = 1$, $\pi_m \in [0, 1]$. 
The first and second terms represent the ELBO on inferences from all modalities. 
In contrast, the third term corresponds to the negative KL divergence between the inferences from all modalities and the unimodal inferences.

In alignment-based multimodal VAEs, we can regard obtaining unimodal inference as a two-step approximation: (1) The true multimodal inference $p_{\theta}(\mathbf{z}|X)$ is approximated by the amortized variational approximation using the neural network $q_{\phi}(\mathbf{z}|X)$ (optimizing the first and second terms in Equation~\eqref{eq:mvtcae}), and (2) The multimodal inference model $q_{\phi}(\mathbf{z}|X)$ is approximated by the unimodal inference model $q_{\lambda_m}(\mathbf{z}|\mathbf{x}_m)$ (optimizing the third terms in Equation~\eqref{eq:mvtcae}). 
Considering that our primary objective is to obtain a unimodal inference $q_{\lambda_m}(\mathbf{z}|\mathbf{x}_m)$ that approximates the true multimodal inference $p_{\theta}(\mathbf{z}|X)$, we can see that whether this goal is achieved depends on how well the multimodal inference model $q_{\phi}(\mathbf{z}|X)$ approximates the true inference.

However, in amortized inferences, there will always be an inference approximation error called the amortization gap, the approximation gap due to variational approximation, between the true multimodal inference \( p_{\theta}(\mathbf{z}|X) \). 
Therefore, in the first-step approximation of alignment-based multimodal VAEs, it becomes difficult to approximate the true inference by unimodal inference.

\subsection{Iterative Amortized Inference}

Amortized variational inference refers to a method of performing variational inference by optimizing shared parameters across the entire dataset rather than optimizing the parameters of the approximate distribution for each data point. 
Amortized variational inference can significantly reduce computational costs when using approximate distributions that are costly to optimize, such as neural networks. However, it is known that this approach can lead to discrepancies with inferences optimized for individual data points, resulting in inferior inference accuracy. 
This discrepancy is referred to as the amortization gap~\citep{cremer18}.
VAEs are also subject to the negative effects of the amortization gap. This is because the inference model is approximated by a neural network \( q_{\phi}(\mathbf{z}|\mathbf{x}) \), sharing parameters \( \phi \) across the entire dataset, and employing amortized variational inference.

Iterative amortized inference is an algorithm that improves inference accuracy by iteratively updating the inference in amortized variational inference, thereby reducing the amortization gap. 
It is known that by applying iterative amortized inference to single-modality VAEs, the reduction in inference accuracy caused by the amortization gap can be improved. 
Iterative amortized inference is used in scenarios with challenging inference, such as object-centric representation learning~\citep{greff19}.

For updating the inference, gradients of the latent variables obtained by backpropagating the error from the VAE's loss function and the differences between the input and reconstruction are used. 
This update is expressed as follows, where the input to the inferrer is \( \mathbf{x} \), the mean of the latent variables at iteration \( t \) is \( {\boldsymbol{\mu}}_{t} \), the standard deviation is \( {\boldsymbol{\sigma}}_{t} \), and the gradients of the mean and standard deviation of the latent variables by ELBO (\( {\mathcal L} \)) are \( \nabla_{\boldsymbol{\mu}}{\mathcal L} \), \( \nabla_{\boldsymbol{\sigma}}{\mathcal L} \), respectively. A parameterized function \( f_{w} \) is used for the update:
\begin{equation}
\label{iai}
\boldsymbol{\mu}_{t+1}, \boldsymbol{\sigma}_{t+1} = f_{w}(\mathbf{x},  \boldsymbol{\mu}_{t}, \boldsymbol{\sigma}_{t}, \nabla_{\boldsymbol{\mu}_{t}}{\mathcal L}, \nabla_{\boldsymbol{\sigma}_{t}}{\mathcal L}).
\end{equation}

\section{Methods}

\begin{figure}[tb]
\centering
\includegraphics[width=\textwidth]{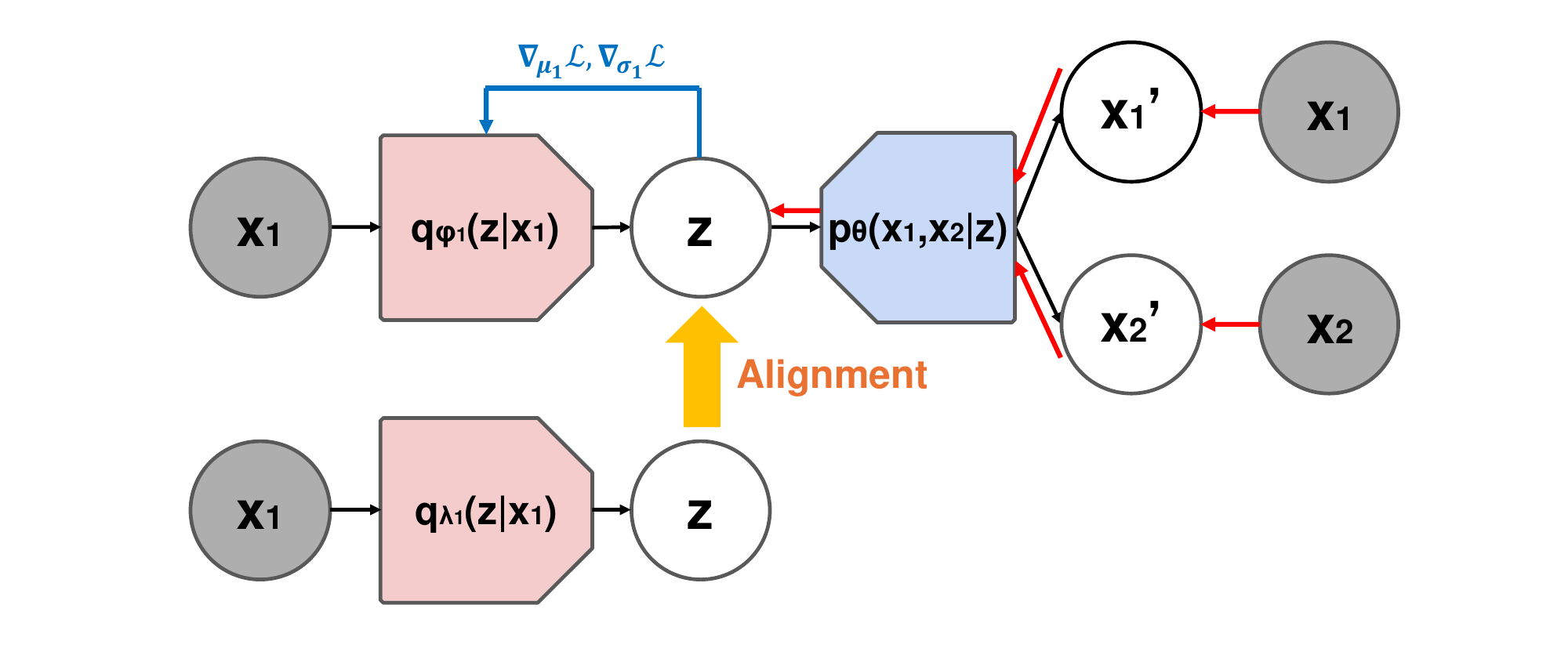}
\caption{Proposed method (\# of modality types $M = 2$). Red lines represent gradient propagation through backpropagation using the multimodal ELBO, while blue lines indicate inference updates using gradients of the mean ($\boldsymbol{\mu}$) and variance ($\boldsymbol{\sigma}$). Together, these form the process of multimodal iterative amortized inference.}
\label{iai_fig}
\end{figure}

To acquire a unimodal inference model \( q_{\lambda_m}(\mathbf{z}|\mathbf{x}_m) \) that approximates the true multimodal inference \( p_{\theta}(\mathbf{z}|X) \), alignment-based multimodal VAEs perform a two-stage approximation. 
This approach avoids the limitations of mixture-based multimodal VAEs, which suffer from information loss due to missing modalities. 
However, in the first stage of approximation, the true multimodal inference \( p_{\theta}(\mathbf{z}|X) \) is approximated by an amortized variational inference model \( q_{\phi}(\mathbf{z}|X) \), which contains errors from the approximation gap of variational inference and the amortization gap inherent in amortized inference.

To acquire a unimodal inference model that closely approximates the true multimodal inference \( p_{\theta}(\mathbf{z}|X) \), it is necessary to reduce these gaps. 
Specifically, our study focuses on reducing the amortization gap, introducing iterative amortized inference in the first stage of approximation. 
By doing so, we aim to improve the approximation accuracy of the multimodal inference and, consequently, enhance the unimodal inference.

However, iterative amortized inference alone cannot perform inference from truly unimodal inputs because the calculation of the multimodal ELBO requires information from all modalities. 
Our goal is to perform inference that incorporates multimodal information using only unimodal inputs. 
To achieve this, we align the unimodal inference \( q_{\lambda_m}(\mathbf{z}|\mathbf{x}_m) \) with the multimodal iterative amortized inference \( q_{\phi_m}(\mathbf{z}_T|\mathbf{x}_m) \).
By minimizing the Kullback–Leibler (KL) divergence between them, we bring the unimodal inference closer to an inference process that overcomes both the information loss due to missing modalities and the amortization gap.

Thus, the proposed method is presented from two perspectives: (1) multimodal iterative amortized inference: improvement of multimodal inference by iterative amortized inference, which overcomes both the information loss due to missing modalities in mixture-based models and the amortization gap in alignment-based models. (2) alignment with unimodal inference: alignment between the multimodal iterative amortized inference and the unimodal inference.

\subsection{Multimodal Iterative Amortized Inference}

Applying iterative amortized inference to multimodal inference improves the amortization gap present in the approximation of the true multimodal inference \( p_{\theta}(\mathbf{z}|X) \) by \( q_{\phi_m}(\mathbf{z}|\mathbf{x}_m) \). 
In our method, we explicitly input only a single modality \( \mathbf{x}_m \), and then we perform iterative amortized inference using all modalities \( X \) during training. 
This approach leverages information from all modalities to refine the inference, even when only a single modality is available as explicit input.

To achieve this, we prepare a generator \( p_{\theta}(X|\mathbf{z}) \) and an iterative amortized inference function \( f_{w} \). 
We can improve the inference by utilizing information from all modalities (see the upper part of Figure~\ref{iai_fig}) by performing iterative amortized inference using the gradient of the multimodal ELBO formulated as follows:
\begin{equation}
\label{loss}
{\mathcal L}({\bf \theta},{\bf \phi};X) = \mathbb{E}_{q(\mathbf{z}_t|\mathbf{x}_m)}[\log p_\theta(X|\mathbf{z}_t)] - D_{\text{KL}}[q_{\phi_m}(\mathbf{z}_t|\mathbf{x}_m)||p(\mathbf{z})].
\end{equation}
In this formulation, even though the inference model \( q_{\phi_m}(\mathbf{z}_t|\mathbf{x}_m) \) takes only \( \mathbf{x}_m \) as input, the reconstruction term \( \log p_\theta(X|\mathbf{z}_t) \) involves all modalities \( X \). 
Therefore, the iterative updates incorporate information from all modalities, allowing us to recover information from missing modalities and reduce the missing modality gap.
Therefore, we not only reduce the amortization gap in the multimodal inference approximated by a neural network but also overcome the information loss due to missing modalities.
The update of multimodal iterative amoritzed inference is formulated as follows:
\begin{equation}
\boldsymbol{\mu}_{m_{t+1}}, \boldsymbol{\sigma}_{m_{t+1}} = f_{w}(\mathbf{x}_m,  \boldsymbol{\mu}_{m_t}, \boldsymbol{\sigma}_{m_t}, \nabla_{\boldsymbol{\mu}_{m_t}}{\mathcal L}, \nabla_{\boldsymbol{\sigma}_{m_t}}{\mathcal L}).
\end{equation}
By repeating this step \( T \) times, we obtain \( \boldsymbol{\mu}_{m_T}, \boldsymbol{\sigma}_{m_T} \), and sample latent variables \( \mathbf{z}_{T} \) from \( q(\mathbf{z}_{T}| \mathbf{x}_{m}) = \mathcal{N}(\boldsymbol{\mu}_{m_T}, \boldsymbol{\sigma}_{m_T}) \), allowing the inference of a shared representation from modality \( m \) that incorporates multimodal information.

\subsection{Alignment of Unimodal Inference with Iterative Amortized Inference}

In multimodal iterative amortized inference, improving inference relies on the multimodal ELBO, which necessitates access to all modalities during computation. 
However, our goal is to perform inference that incorporates multimodal information using only unimodal inputs. 
To achieve this, we align the unimodal inference \( q_{\lambda_m}(\mathbf{z}|\mathbf{x}_m) \) with the multimodal iterative amortized inference, thereby obtaining the desired unimodal inference (see the lower part of Figure~\ref{iai_fig}). 
Our ultimate goal is to acquire a unimodal inference model  that closely approximates the true multimodal inference \( p_{\theta}(\mathbf{z}|X) \). 

By minimizing the KL divergence between the unimodal inference \( q_{\lambda_m}(\mathbf{z}|\mathbf{x}_m) \) and the multimodal iterative amortized inference \( q_{\phi_m}(\mathbf{z}_{T}|\mathbf{x}_{m}) \), we bring the unimodal inference closer to an inference process that overcomes both the information loss due to missing modalities and the amortization gap.

The alignment is performed by minimizing the following KL divergence:

\begin{equation}
\label{distillation}
 {\mathcal D}(\lambda;X) = \sum_{m=1}^{M} D_{\text{KL}}[q_{\phi_m}(\mathbf{z}_T|\mathbf{x}_{m})||q_{\lambda_m}(\mathbf{z}|\mathbf{x}_m)],
\end{equation}
This corresponds to the alignment step in alignment-based multimodal VAEs (the third term in Equation~\ref{eq:mvtcae}), but with the key difference that the source of alignment, \( q_{\phi_m}(\mathbf{z}_T|\mathbf{x}_{m}) \), is obtained through iterative amortized inference using all modalities, even though it takes only a single modality as input. 
By aligning the unimodal inference to such a multimodal iterative amortized posterior, we obtain unimodal inferences that are sourced from an inference process that overcomes both the information loss due to missing modalities and the amortization gap.

\section{Related Works}
Research on multimodal VAEs is being conducted in various ways beyond what was mentioned above~\citep{suzuki22}. Early multimodal VAEs did not devise Aggregation methods like those seen in Mixed-based. Hence, an exponential number of models had to be prepared for the number of modalities. JMVAE~\citep{suzuki16} prepares inferences from two modalities and a single modality and learns unimodal inference by bringing them closer together. This method is equivalent to MVTCAE when the number of modalities is two, but MVTCAE differs in that it uses the PoE of unimodal inference for multimodal inference. Furthermore, TELBO~\citep{vedantam18} is proposed as the sum of ELBO for all combinations of modalities, and as further studies, M\(^2\)VAE~\citep{korthals19} combining JMVAE and TELBO, and VAEVAE~\citep{wu19}, which excludes the term of KL divergence between multimodal inference and prior distribution, have been proposed.

In addition, some studies introduce modality-specific latent variables~\citep{tsai18, hsu18, sutter20, lee20, daunhawer21b, palumbo23} or hierarchical latent variables~\citep{sutter21B, wolff21, vasco22}. Since our study aims to acquire shared representations of all modalities from a single modality, these studies complement ours.

Recent models dealing with multimodal information include those that use large architectures (e.g., Transformer~\citep{vaswani17}) and a large amount of data. Conditionally generated models are actively being studied, and, for example, DALL-E2~\citep{ramesh22} and Imagen~\citep{saharia22} enable high-quality image generation conditioned on language. However, these models focus on generation rather than multimodal representation learning. Studies on representation learning from multimodal information include CLIP~\citep{radford21} and MultiMAE~\citep{bachmann22}. In CLIP, images and languages are encoded with their respective encoders and then learned to be close in the representation space through contrastive learning. MultiMAE is conducting representation learning and cross-modal generation against diverse image information with a large-scale model based on the Vision Transformer~\citep{dosovitskiy20}, which allows cross-modal generation. Still, it is not aimed at acquiring shared representations of modalities.

\section{Experiments}

In this study, we used the MoPoE-VAE~\citep{sutter21} as the mixture-based model, employing MoPoE for combining the multimodal distributions. 
For the alignment-based model, based on prior work~\citep{hwang21}, we used a VAE that integrates multimodal information using the Product of Experts (PoE) as the alignment source model. 
Moreover, since we wanted to separately examine the properties of both the alignment source and the target unimodal inference in our proposed method, we conducted training in two stages: first learning the alignment source distribution, and then learning the acquisition of unimodal inference through distribution alignment.

For the network architecture, we followed~\citep{sutter20} and~\citep{daunhawer21}, setting the dimension of the latent variable to 1024 based on the dimensions used for SVHN in the iterative amortized inference paper~\citep{marino18}. 
This architectural setting was shared across all experiments.
Implementation details of multimodal iterative amoritzed inference are shown in Appendix~\ref{implementation_details}.
For all models, we used the Adam optimizer~\citep{kingma14} with a learning rate of $0.0002$. 
Additionally, we employed a learning rate scheduler using ExponentialLR with a gamma of $0.98$. 
The batch size was set to 256 for MNIST-SVHN-Text and 128 for CUB. 
Training was performed for 100 epochs on MNIST-SVHN-Text and 200 epochs on CUB.

We conducted experiments using the widely used multimodal datasets MNIST-SVHN-Text~\citep{sutter20}, composed of three modalities, and CUB~\citep{wah2011caltech}, consisting of two modalities. 
MNIST-SVHN-Text is a dataset related to digits from 0 to 9, and the three constituting modalities are MNIST, a $28 \times 28$ grayscale image of digits; SVHN, a $3 \times 32 \times 32$ RGB image of digits; and Text, the alphabetical notation of numbers randomly placed within an 8-word frame. 
CUB is a dataset with bird images and their captions. 
CUB images were compressed to $3 \times 64 \times 64$.
Each dataset was split into training, validation, and test sets. 
We used the validation set to monitor the training process and select models. 
For the second stage of training in alignment-based models and our proposed method, the number of epochs remained the same as in the first stage. 
However, we selected the model where the KL divergence between the alignment source and the target unimodal inference on the validation data was minimized.

\subsection{Effectiveness of Multimodal Iterative Amortized Inference}

\begin{figure}[t]
\centering
\begin{minipage}[b]{0.49\columnwidth}
    \centering
    \includegraphics[width=1.0\columnwidth]{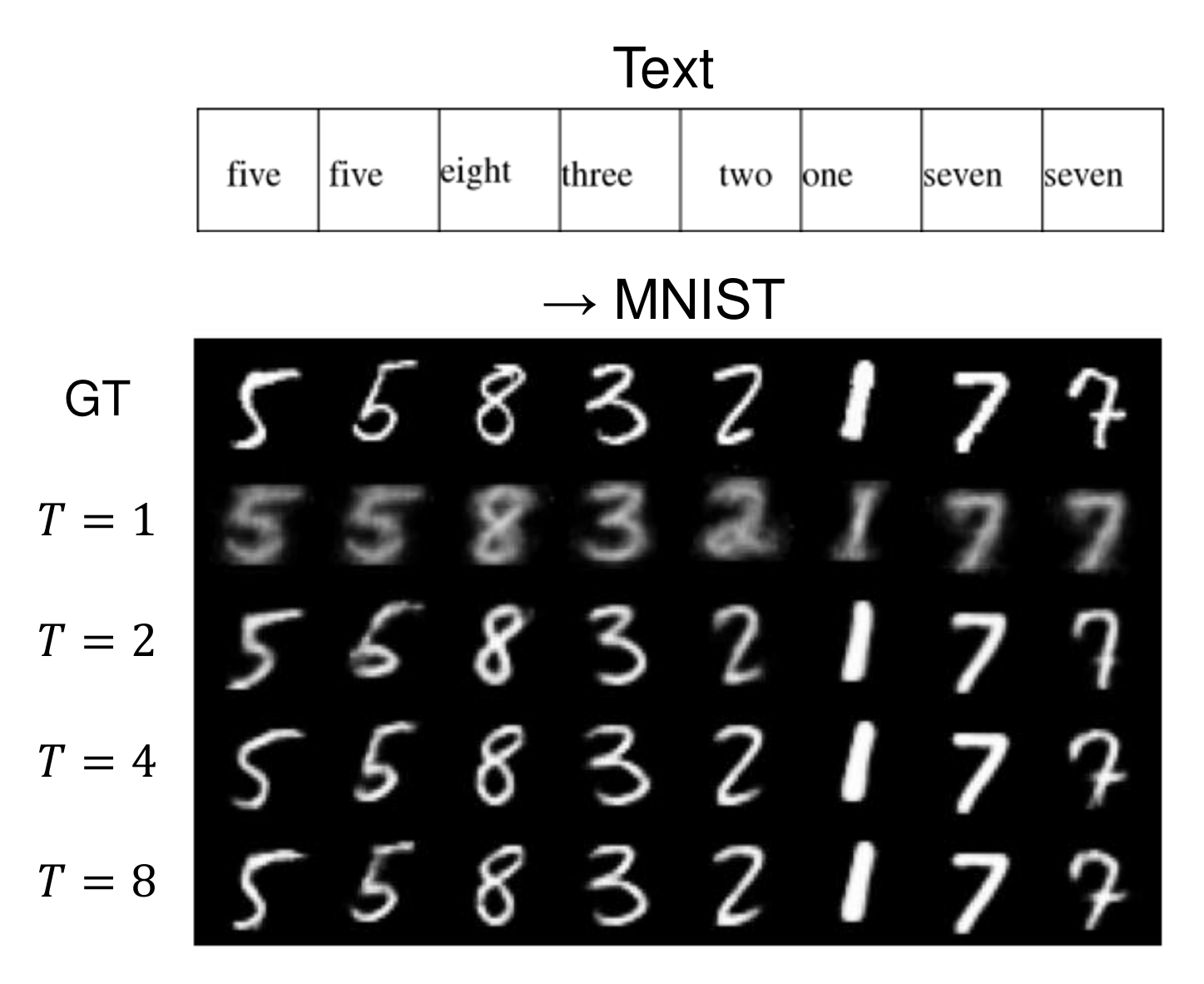}
\end{minipage}
\begin{minipage}[b]{0.49\columnwidth}
    \centering
    \includegraphics[width=1.0\columnwidth]{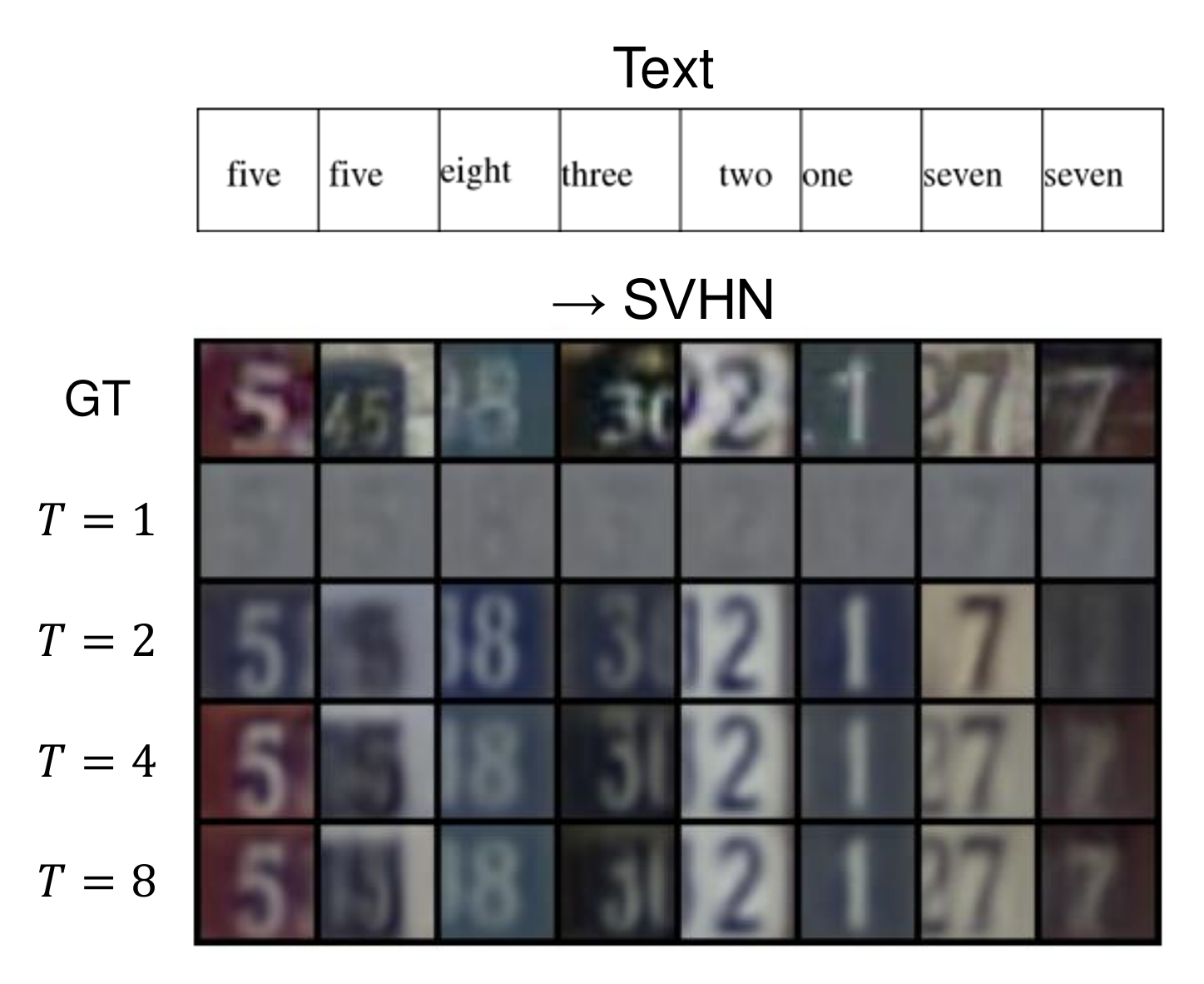}
\end{minipage}
\caption{Qualitative results of cross-modal generation on the MNIST-SVHN-Text dataset when applying multimodal iterative amortized inference to \( q_{\phi_2}(\mathbf{z}|\mathbf{x}_2) \) (input modality $\mathbf{x}_2$ is Text, generated modality $\mathbf{x}_0$ is MNIST (left), and generated modality $\mathbf{x}_1$ is SVHN (right)). By increasing the number of iterations \( T \), information from missing modalities is recovered, improving the performance of cross-modal generation.}
\label{iterative_generation}  
\end{figure}

\begin{figure}[t]
  \centering
  \includegraphics[width=\textwidth]{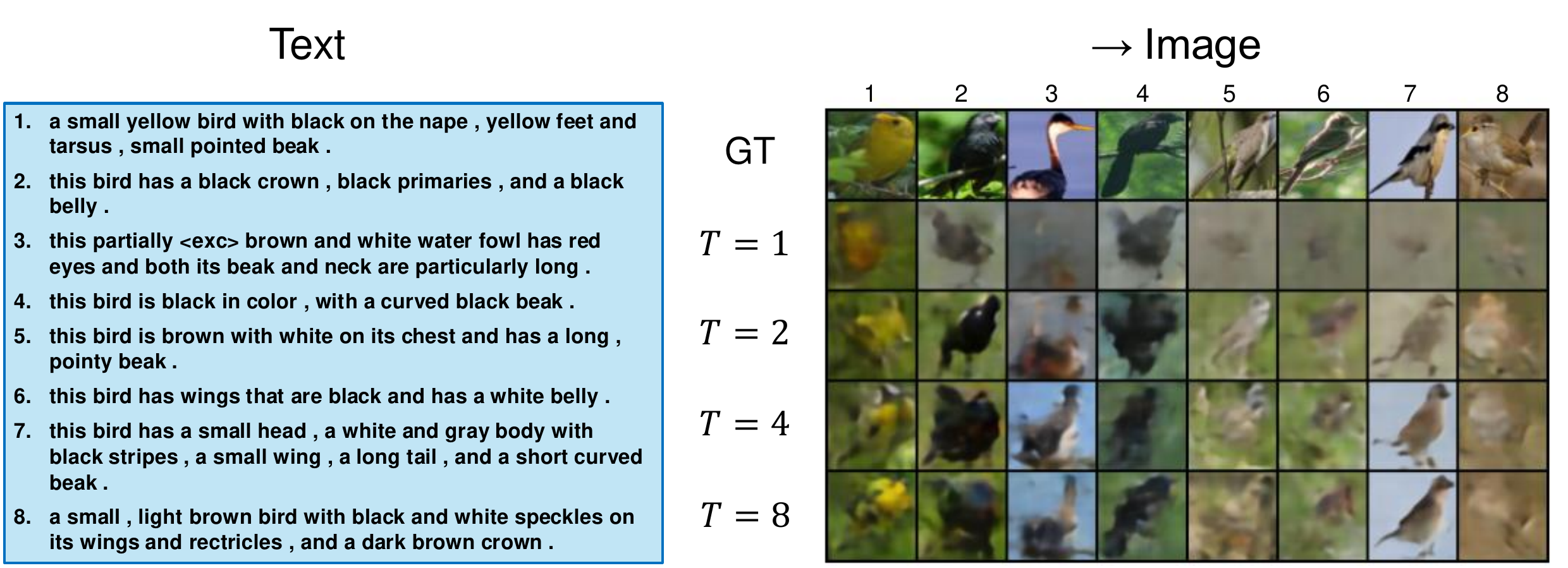}
  \caption{Qualitative results of cross-modal generation on the CUB dataset when applying multimodal iterative amortized inference to \( q_{\phi_1}(\mathbf{z}|\mathbf{x}_1) \) (input modality $\mathbf{x}_1$ is Text (shown in a blue box), generated modality $\mathbf{x}_0$ is Image. By increasing the number of iterations \( T \), information from missing modalities is recovered, improving the performance of cross-modal generation.}
  \label{iterative_generation_cub}
\end{figure}

To demonstrate that multimodal iterative amortized inference can compensate for missing modality information and reduce the amortization gap, we conducted experiments using the original undistilled model.
For both datasets, the number of iteration $T = 8$ in training time of the proposed method.

First, we present qualitative results of cross-modal generation to demonstrate how our proposed method addresses the information loss due to missing modalities inherent in mixture-based models.
We apply multimodal iterative amortized inference to the unimodal inference model \( q_{\phi_m}(\mathbf{z}|\mathbf{x}_m) \), where the input is unimodal. 
Figure~\ref{iterative_generation} and Figure~\ref{iterative_generation_cub} shows generated images from text inputs with different numbers of iterations \( T = 1, 2, 4, 8 \). 
When \( T = 1 \) (i.e., no iterative updates), the cross-modal generation may fail to capture detailed information from the missing image modality, resulting in poor-quality images. 
However, as we increase \( T \), the quality of the generated images improves significantly. 
The iterative amortized inference effectively recovers information from the missing modalities by leveraging the multimodal ELBO.
This demonstrates that our method overcomes the information loss due to missing modalities, a limitation in mixture-based models.
More qualitative results are shown in ~\ref{teacher-qualitative-appendix}.

\begin{figure}[t]
\centering
\includegraphics[width=\textwidth]{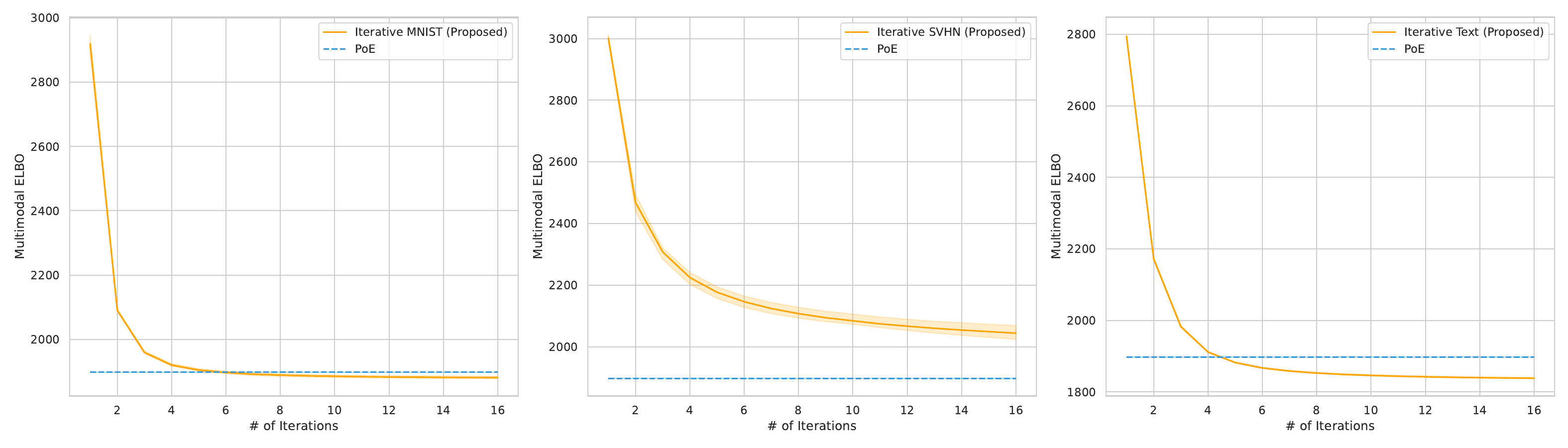}
\caption{Improvement in multimodal ELBO using multimodal iterative amortized inference to \( q_{\phi_0}(\mathbf{z}|\mathbf{x}_0) \) (input modality $\mathbf{x}_0$ is MNIST, left), to \( q_{\phi_1}(\mathbf{z}|\mathbf{x}_1) \) (input modality $\mathbf{x}_1$ is SVHN, middle) and to \( q_{\phi_2}(\mathbf{z}|\mathbf{x}_2) \) (input modality $\mathbf{x}_2$ is Text, right) on the MNIST-SVHN-Text dataset. Image The dotted line represents the ELBO of the alignment source model (PoE).}
\label{iteration_elbo}
\end{figure}

\begin{figure}[t]
\centering
\includegraphics[width=\textwidth]{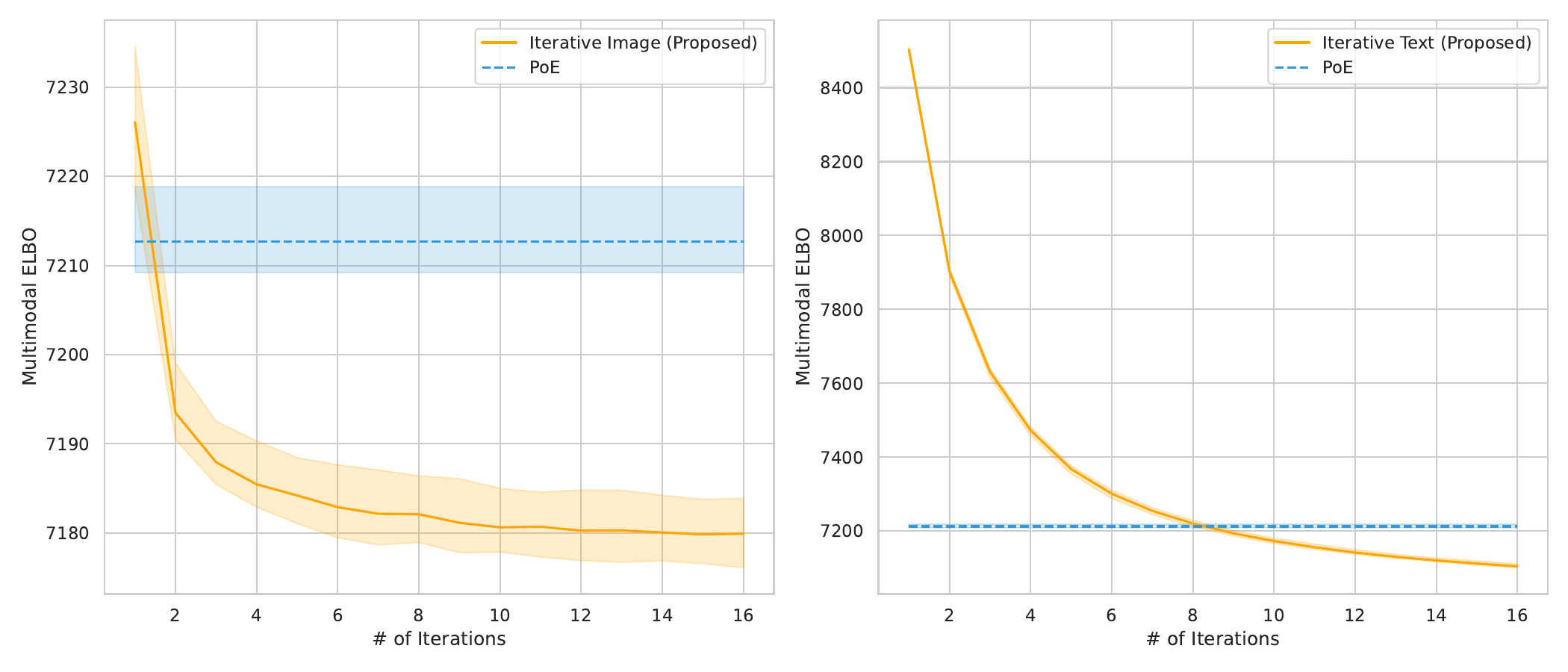}
\caption{Improvement in multimodal ELBO using multimodal iterative amortized inference to \( q_{\phi_0}(\mathbf{z}|\mathbf{x}_0) \) (input modality $\mathbf{x}_0$ is Image, left) and to \( q_{\phi_1}(\mathbf{z}|\mathbf{x}_1) \) (input modality $\mathbf{x}_1$ is Text, right) on the CUB dataset. Image The dotted line represents the ELBO of the alignment source model (PoE).}
\label{iteration_elbo_cub}
\end{figure}

Next, we analyze the improvement in the evidence lower bound (ELBO) as we increase the number of iterations \( T \), highlighting how our method reduces the amortization gap present in alignment-based models.
Figure~\ref{iteration_elbo} and Figure~\ref{iteration_elbo_cub} shows the relationship between the ELBO and the number of iterations from \( T = 1 \) to \( T = 16 \). 
For comparison, we also plot the ELBO values for the Product of Experts (PoE) model, which serves as the alignment source model in alignment-based models in this study, with dotted lines.
The PoE model combines unimodal inferences to approximate the multimodal posterior but suffers from the amortization gap due to the limitations of amortized inference.
As shown in Figure~\ref{iteration_elbo} and Figure~\ref{iteration_elbo_cub}, increasing the number of iterations \( T \) leads to improved ELBO values that surpass those of the alignment source model (PoE) in most cases. 
This indicates that our iterative amortized inference method effectively reduces the amortization gap inherent in the alignment-based models. 
By refining the inference through iterative updates, we achieve a better approximation of the true multimodal posterior than the PoE model, resulting in improved inference performance.

In summary, the qualitative results of cross-modal generation demonstrate that our method overcomes the information loss due to missing modalities in mixture-based models. 
The ELBO improvements with increasing \( T \) show that our method reduces the amortization gap present in alignment-based models by outperforming the alignment source model (PoE). 
These findings confirm the effectiveness of our proposed multimodal iterative amortized inference in addressing both key challenges.

\subsection{Aligning Unimodal Inference with Iterative Amortized Inference}

In multimodal iterative amortized inference, improving the inference relies on the multimodal ELBO, which requires access to all modalities during computation. 
However, our goal is to perform inference that incorporates multimodal information using only unimodal inputs. To achieve this, we align the unimodal inference \( q_{\lambda_m}(\mathbf{z}|\mathbf{x}_m) \) with the multimodal iterative amortized inference, thereby obtaining a unimodal inference model that closely approximates the true multimodal inference. 
We set the number of iterations \( T = 8 \) for all settings and evaluate the effectiveness of our approach by assessing both the performance of the inferred latent representations and the quality of cross-modal generation.

To evaluate the performance of the inferred latent representations and the cross-modal generation quality, we conducted experiments on the MNIST-SVHN-Text dataset using two evaluation metrics: linear classification accuracy and FID~\citep{heusel2017gans}. 
The linear classification accuracy measures the quality of the latent representations themselves, indicating how well they capture the underlying data structure. 
The FID is a metric that measures the distance between the distribution of generated samples from a generative model and the distribution of the dataset. 
By using FID in the evaluation of cross-modal generation, we can assess how much the unimodal inference includes information from other modalities.

Table~\ref{tab:linear-mst} presents the linear classification accuracy of the latent variables obtained from different inference models. 
Our proposed method outperforms alignment-based models and surpasses mixture-based models in most settings. 
This indicates that our method effectively captures the underlying data structure in the latent space.
Table~\ref{tab:fid-mst} shows the FID for cross-modal generation. Lower FID indicate better quality of the generated samples.
Our method significantly outperforms mixture-based models and outperforms alignment-based models in most settings. 
This demonstrates that our approach enables the cross-modal generation to effectively capture the data distribution of the other modalities, resulting in higher quality generated samples.

For the CUB dataset, we evaluated the models using two metrics: the FID of cross-modal generation and the cosine similarity of representations inferred from the two modalities. 
The cosine similarity assesses the quality of the latent representations themselves by measuring the similarity between representations inferred from different modalities.
As shown in Table~\ref{tab:cub-quantitative}, our proposed method significantly outperforms mixture-based models in terms of FID and also outperforms alignment-based models. 
For cosine similarity, our method significantly outperforms mixture-based models and achieves competitive results compared to alignment-based models.

\begin{figure}[tb]
\centering
\begin{minipage}[b]{0.49\columnwidth}
    \centering
    \includegraphics[width=1.0\columnwidth]{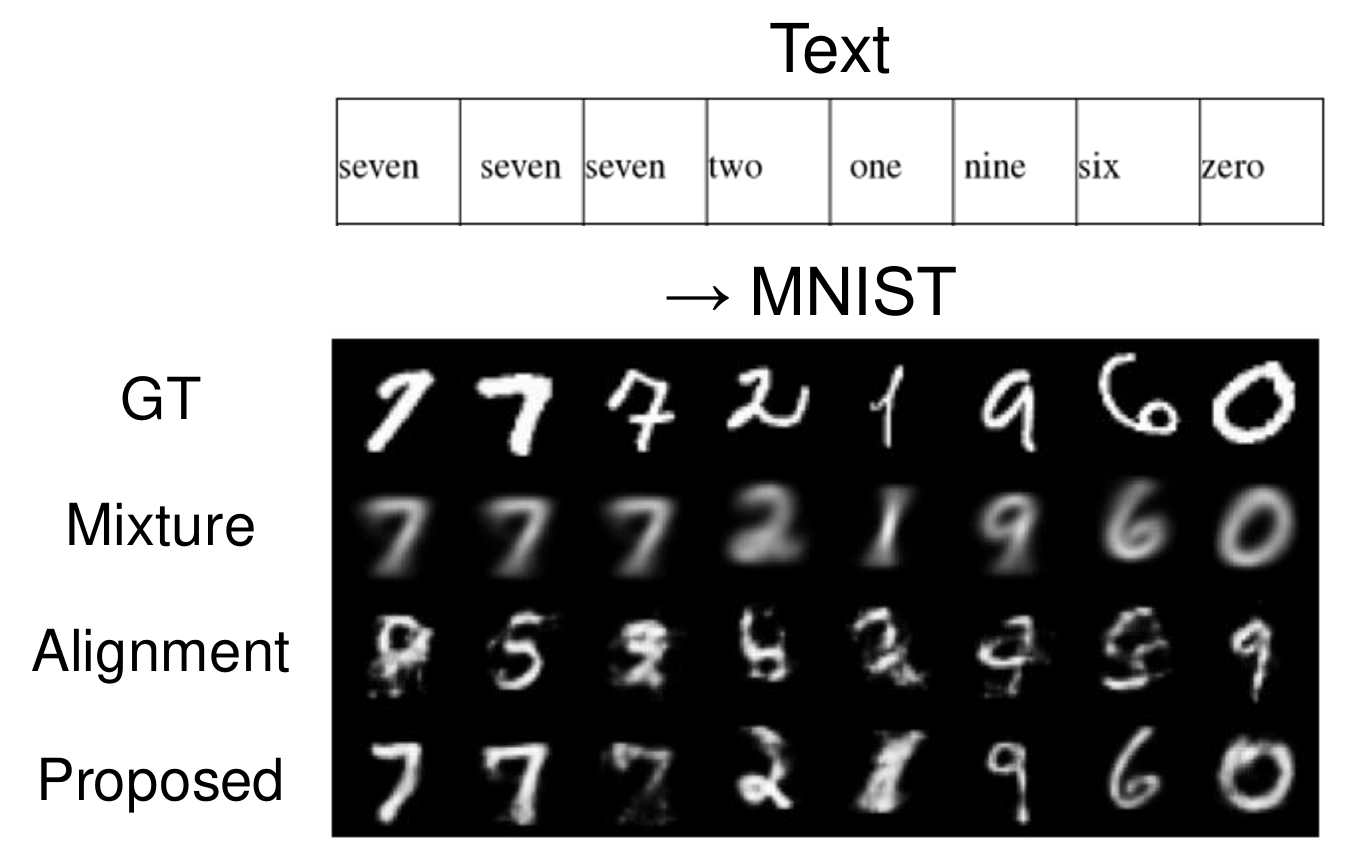}
\end{minipage}
\begin{minipage}[b]{0.49\columnwidth}
    \centering
    \includegraphics[width=1.0\columnwidth]{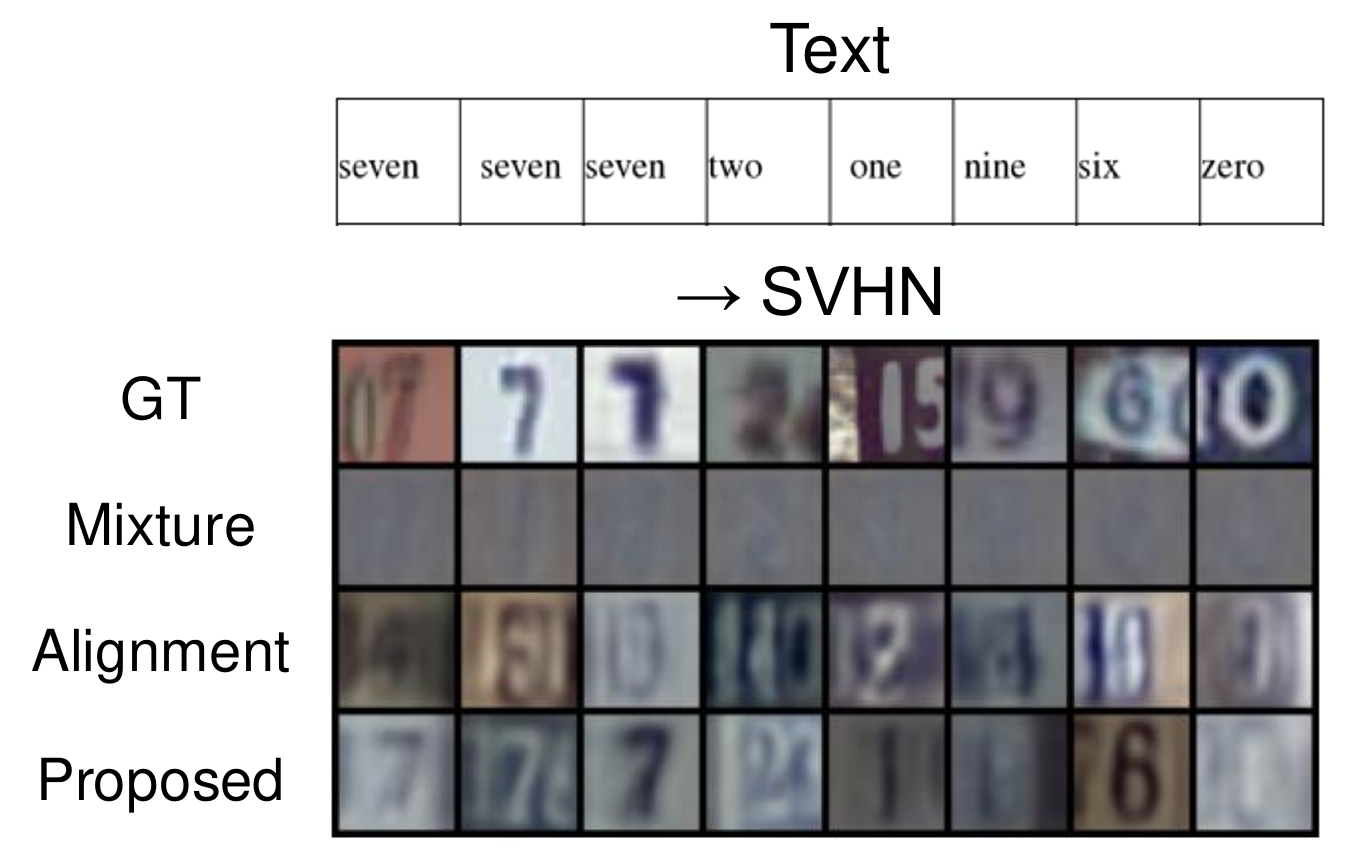}
\end{minipage}
\caption{Comparison of cross-modal generation on the MNIST-SVHN-Text dataset (input modality $\mathbf{x}_2$ is Text, generated modality $\mathbf{x}_0$ is MNIST (left), and generated modality $\mathbf{x}_1$ is SVHN (right)). }
\label{comparison_student_2}  
\end{figure}

\begin{figure}[tb]
\centering
\includegraphics[width=\textwidth]{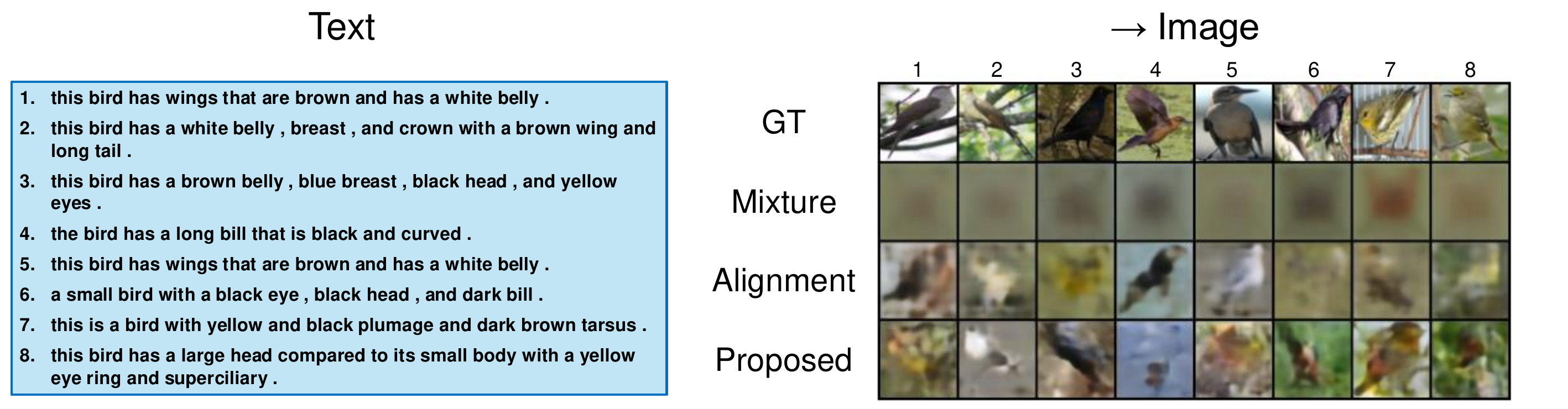}
\caption{Comparison of cross-modal generation on the CUB dataset (input modality $\mathbf{x}_1$ is Text (left), generated modality $\mathbf{x}_0$ is Image (right)). }
\label{comparison_student_cub}  
\end{figure}

Figures~\ref{comparison_student_2} and \ref{comparison_student_cub} illustrate the quality of cross-modal generation for different models.
First, consistent with previous research~\citep{daunhawer21}, we confirmed that mixture-based models are unable to perform cross-modal generation effectively.
Next, the performance difference in cross-modal generation between alignment-based models and our proposed method is qualitatively observed as shown in these figures.

\begin{table*}
\centering
\caption{Linear classification accuracy (\%) by latent representation on MNIST-SVHN-Text dataset.}
\begin{tabular}{l||ccc}
\hline
Model & MNIST & SVHN & Text \\
\hline
Mixture-based & 97.99 & \textbf{78.92} & 99.78 \\
Alignment-based & 96.72 & 76.49 & 99.75 \\
Proposed & \textbf{98.15} & \underline{78.50} & \textbf{99.97} \\
\hline
\end{tabular}
\label{tab:linear-mst}
\end{table*}

\begin{table*}
\centering
\caption{FID for cross-modal geneartion on MNIST-SVHN-Text dataset. Lower is better. M, S, and T stand for MNIST, SVHN, and Text.}
\begin{tabular}{l||cccc}
\hline
Model & S $\to$ M & T $\to$ M & M $\to$ S & T $\to$ S \\
\hline
Mixture-based & 201.49 & 209.53 & 214.05 & 215.79 \\
Alignment-based & 60.90 & 48.98 & \textbf{49.23} & 49.74 \\
Proposed & \textbf{50.13} & \textbf{40.27} & \underline{49.43} & \textbf{48.39} \\
\hline
\end{tabular}
\label{tab:fid-mst}
\end{table*}

\begin{table*}[t]
\caption{Qualitative results on CUB dataset. (left) FID for cross-modal generation on CUB dataset. Lower is better. (right) Cosine similarity of latent representation from image and text on CUB dataset. Higher is better.}
\centering
\begin{tabular}{ccc}
    \centering
    \begin{tabular}{l||c|c}
    \hline
    Model & FID (Text $\to$ Image) & Cosine Similarity \\
    \hline
    Mixture-based & 325.98 & $7.909 \times 10^{-4}$ \\
    Alignment-based & 268.15 & $\bf{2.963 \times 10^{-2}}$ \\
    Proposed & \textbf{207.13} & \underline{$2.862 \times 10^{-2}$} \\
    \hline
    \end{tabular}
    \label{tab:cub-quantitative}
\end{tabular}
\end{table*}

\section{Limitation and Conclusion}

This study found that multimodal iterative amortized inference can improve not only the amortization gap but also the missing modality information loss caused by modality sub-sampling. 
By aligning multimodal iterative amortized inference with unimodal inference, we were able to obtain an improved unimodal inference model. 
The results demonstrated that our proposed method improves the performance of the inference itself, as evidenced by higher linear classification accuracy and greater cosine similarity of latent representations. 
Additionally, the representations learned by our method effectively capture the distributions of other modalities, which is reflected in favorable FID in cross-modal generation. 

However, the proposed method has some limitations. 
To assess the consistency of cross-modal generation, we measured the classification accuracy of pre-trained classifiers~\citep{sutter21} on the MNIST-SVHN-Text dataset (Table~\ref{tab:classifier-mst}). 
While our method achieved high-quality generation in terms of distribution alignment, the classification results revealed that numerical consistency was not fully maintained in some cases. 
This suggests that iterative amortized inference is proficient at modeling the distributions of missing modalities, but there may be a trade-off with maintaining exact content consistency.

Overall, our approach significantly improves over mixture-based models and offers competitive or even better performance compared to alignment-based models. 
In future work, further enhancements in the accuracy of the obtained unimodal inference can be achieved by refining the coordination method between multimodal iterative amortized inference and unimodal inference. 
Developing a coordination method that is more compatible with iterative updates and performs alignment more effectively than minimizing the KL divergence once per learning step may lead to even better performance.

\begin{table*}
\centering
\caption{Cross-coherence by classification accuracy (\%) on MNIST-SVHN-Text dataset. M, S, and T stand for MNIST, SVHN, and Text.}
\begin{tabular}{l||cccccc}
\hline
Model & S $\to$ M & T $\to$ M & M $\to$ S & T $\to$ S & M $\to$ T & S $\to$ T \\
\hline
Mixture-based & \textbf{77.19} & \textbf{98.61} & 26.88 & 27.02 & \textbf{97.95} & \textbf{78.59} \\
Alignment-based & 66.94 & 92.39 & 75.36 & 77.72 & 94.46 & 73.35 \\
Proposed & \underline{75.63} & \underline{98.41} & \textbf{82.11} & \textbf{84.90} & \underline{97.70} & \underline{77.33} \\
\hline
\end{tabular}
\label{tab:classifier-mst}
\end{table*}

\section*{}
\textbf{Availability Statement}~The experiment code for this study is not publicly available. The datasets
used for the experiments in this study are publicly available through the Internet.

\noindent
\textbf{Funding}~This work was supported by the Japan Society for the Promotion of Science (JSPS) KAKENHI Grant Number J23H04974. 

\section*{Declarations}
\textbf{Confict of interest}~The authors declare no competing interests relevant to the contents of this article.

\bibliography{iai}

\begin{thebibliography}{10}
\expandafter\ifx\csname url\endcsname\relax
  \def\url#1{\burl{#1}}\fi
\expandafter\ifx\csname urlprefix\endcsname\relax\def\urlprefix{URL }\fi
\providecommand{\bibinfo}[2]{#2}
\providecommand{\eprint}[2][]{\url{#2}}
\providecommand{\doi}[1]{\url{https://doi.org/#1}}
\bibcommenthead

\bibitem{baltrusaitis18}
\bibinfo{author}{Baltru{\v{s}}aitis, T.}, \bibinfo{author}{Ahuja, C.} \& \bibinfo{author}{Morency, L.-P.}
\newblock \bibinfo{title}{Multimodal machine learning: A survey and taxonomy}.
\newblock \emph{\bibinfo{journal}{IEEE Transactions on Pattern Analysis and Machine Intelligence}} \textbf{\bibinfo{volume}{41}}, \bibinfo{pages}{423--443} (\bibinfo{year}{2018}).

\bibitem{kingma13}
\bibinfo{author}{Kingma, D.~P.} \& \bibinfo{author}{Welling, M.}
\newblock \bibinfo{title}{Auto-encoding variational bayes}.
\newblock \emph{\bibinfo{journal}{arXiv preprint arXiv:1312.6114}}  (\bibinfo{year}{2013}).

\bibitem{suzuki16}
\bibinfo{author}{Suzuki, M.}, \bibinfo{author}{Nakayama, K.} \& \bibinfo{author}{Matsuo, Y.}
\newblock \bibinfo{title}{Joint multimodal learning with deep generative models}.
\newblock \emph{\bibinfo{journal}{arXiv preprint arXiv:1611.01891}}  (\bibinfo{year}{2016}).

\bibitem{suzuki22}
\bibinfo{author}{Suzuki, M.} \& \bibinfo{author}{Matsuo, Y.}
\newblock \bibinfo{title}{A survey of multimodal deep generative models}.
\newblock \emph{\bibinfo{journal}{Advanced Robotics}} \textbf{\bibinfo{volume}{36}}, \bibinfo{pages}{1019--1026} (\bibinfo{year}{2022}).

\bibitem{wu18}
\bibinfo{author}{Wu, M.} \& \bibinfo{author}{Goodman, N.}
\newblock \bibinfo{title}{Multimodal generative models for scalable weakly-supervised learning}.
\newblock \emph{\bibinfo{journal}{Advances in Neural Information Processing Systems}} \bibinfo{pages}{5575--5585} (\bibinfo{year}{2018}).

\bibitem{shi19}
\bibinfo{author}{Shi, Y.}, \bibinfo{author}{Siddharth, N.}, \bibinfo{author}{Paige, B.} \& \bibinfo{author}{Torr, P.}
\newblock \bibinfo{title}{Variational mixture-of-experts autoencoders for multi-modal deep generative models}.
\newblock \emph{\bibinfo{journal}{Advances in Neural Information Processing Systems}} \bibinfo{pages}{15718--15729} (\bibinfo{year}{2019}).

\bibitem{sutter21}
\bibinfo{author}{Sutter, T.~M.}, \bibinfo{author}{Daunhawer, I.} \& \bibinfo{author}{Vogt, J.~E.}
\newblock \bibinfo{title}{Generalized multimodal elbo}.
\newblock \emph{\bibinfo{journal}{arXiv preprint arXiv:2105.02470}}  (\bibinfo{year}{2021}).

\bibitem{hinton02}
\bibinfo{author}{Hinton, G.~E.}
\newblock \bibinfo{title}{Training products of experts by minimizing contrastive divergence}.
\newblock \emph{\bibinfo{journal}{Neural Computation}} \textbf{\bibinfo{volume}{14}}, \bibinfo{pages}{1771--1800} (\bibinfo{year}{2002}).

\bibitem{daunhawer21}
\bibinfo{author}{Daunhawer, I.}, \bibinfo{author}{Sutter, T.~M.}, \bibinfo{author}{Chin-Cheong, K.}, \bibinfo{author}{Palumbo, E.} \& \bibinfo{author}{Vogt, J.~E.}
\newblock \bibinfo{title}{On the limitations of multimodal vaes}.
\newblock \emph{\bibinfo{journal}{arXiv preprint arXiv:2110.04121}}  (\bibinfo{year}{2021}).

\bibitem{hwang21}
\bibinfo{author}{Hwang, H.} \emph{et~al.}
\newblock \bibinfo{title}{Multi-view representation learning via total correlation objective}.
\newblock \emph{\bibinfo{journal}{Advances in Neural Information Processing Systems}} \textbf{\bibinfo{volume}{34}}, \bibinfo{pages}{12194--12207} (\bibinfo{year}{2021}).

\bibitem{cremer18}
\bibinfo{author}{Cremer, C.}, \bibinfo{author}{Li, X.} \& \bibinfo{author}{Duvenaud, D.}
\newblock \bibinfo{title}{Inference suboptimality in variational autoencoders}.
\newblock \emph{\bibinfo{journal}{International Conference on Machine Learning}}  (\bibinfo{year}{2018}).

\bibitem{sutter20}
\bibinfo{author}{Sutter, T.~M.}, \bibinfo{author}{Daunhawer, I.} \& \bibinfo{author}{Vogt, J.~E.}
\newblock \bibinfo{title}{Multimodal generative learning utilizing jensen-shannon divergence}.
\newblock \emph{\bibinfo{journal}{arXiv preprint arXiv:2006.08242}}  (\bibinfo{year}{2020}).

\bibitem{wah2011caltech}
\bibinfo{author}{Wah, C.}, \bibinfo{author}{Branson, S.}, \bibinfo{author}{Welinder, P.}, \bibinfo{author}{Perona, P.} \& \bibinfo{author}{Belongie, S.}
\newblock \bibinfo{title}{The caltech-ucsd birds-200-2011 dataset}.
\newblock \bibinfo{type}{Tech. Rep.}, \bibinfo{institution}{California Institute of Technology} (\bibinfo{year}{2011}).

\bibitem{heusel17}
\bibinfo{author}{Heusel, M.}, \bibinfo{author}{Ramsauer, H.}, \bibinfo{author}{Unterthiner, T.}, \bibinfo{author}{Nessler, B.} \& \bibinfo{author}{Hochreiter, S.}
\newblock \bibinfo{title}{Gans trained by a two time-scale update rule converge to a local nash equilibrium}.
\newblock \emph{\bibinfo{journal}{Advances in Neural Information Processing Systems}} \textbf{\bibinfo{volume}{30}} (\bibinfo{year}{2017}).

\bibitem{greff19}
\bibinfo{author}{Greff, K.} \emph{et~al.}
\newblock \bibinfo{title}{Multi-object representation learning with iterative variational inference}.
\newblock \emph{\bibinfo{journal}{International Conference on Machine Learning}}  (\bibinfo{year}{2019}).

\bibitem{vedantam18}
\bibinfo{author}{Vedantam, R.}, \bibinfo{author}{Fischer, I.}, \bibinfo{author}{Huang, J.} \& \bibinfo{author}{Murphy, K.}
\newblock \bibinfo{title}{Generative models of visually grounded imagination}.
\newblock \emph{\bibinfo{journal}{International Conference on Learning Representations}}  (\bibinfo{year}{2018}).

\bibitem{korthals19}
\bibinfo{author}{Korthals, T.}, \bibinfo{author}{Rudolph, D.}, \bibinfo{author}{Leitner, J.}, \bibinfo{author}{Hesse, M.} \& \bibinfo{author}{R{\"u}ckert, U.}
\newblock \bibinfo{title}{Multi-modal generative models for learning epistemic active sensing}.
\newblock \emph{\bibinfo{journal}{2019 International Conference on Robotics and Automation (ICRA)}} \bibinfo{pages}{3319--3325} (\bibinfo{year}{2019}).

\bibitem{wu19}
\bibinfo{author}{Wu, M.} \& \bibinfo{author}{Goodman, N.}
\newblock \bibinfo{title}{Multimodal generative models for compositional representation learning}.
\newblock \emph{\bibinfo{journal}{arXiv preprint arXiv:1912.05075}}  (\bibinfo{year}{2019}).

\bibitem{tsai18}
\bibinfo{author}{Tsai, Y.-H.~H.}, \bibinfo{author}{Liang, P.~P.}, \bibinfo{author}{Zadeh, A.}, \bibinfo{author}{Morency, L.-P.} \& \bibinfo{author}{Salakhutdinov, R.}
\newblock \bibinfo{title}{Learning factorized multimodal representations}.
\newblock \emph{\bibinfo{journal}{arXiv preprint arXiv:1806.06176}}  (\bibinfo{year}{2018}).

\bibitem{hsu18}
\bibinfo{author}{Hsu, W.-N.} \& \bibinfo{author}{Glass, J.}
\newblock \bibinfo{title}{Disentangling by partitioning: A representation learning framework for multimodal sensory data}.
\newblock \emph{\bibinfo{journal}{arXiv preprint arXiv:1805.11264}}  (\bibinfo{year}{2018}).

\bibitem{lee20}
\bibinfo{author}{Lee, M.} \& \bibinfo{author}{Pavlovic, V.}
\newblock \bibinfo{title}{Private-shared disentangled multimodal vae for learning of hybrid latent representations}.
\newblock \emph{\bibinfo{journal}{arXiv preprint arXiv:2012.13024}}  (\bibinfo{year}{2020}).

\bibitem{daunhawer21b}
\bibinfo{author}{Daunhawer, I.}, \bibinfo{author}{Sutter, T.~M.}, \bibinfo{author}{Marcinkevi{\v{c}}s, R.} \& \bibinfo{author}{Vogt, J.~E.}
\newblock \bibinfo{title}{Self-supervised disentanglement of modality-specific and shared factors improves multimodal generative models}.
\newblock \emph{\bibinfo{journal}{Pattern Recognition}} \textbf{\bibinfo{volume}{12544}}, \bibinfo{pages}{459} (\bibinfo{year}{2021}).

\bibitem{palumbo23}
\bibinfo{author}{Palumbo, E.}, \bibinfo{author}{Daunhawer, I.} \& \bibinfo{author}{Vogt, J.~E.}
\newblock \bibinfo{title}{Mmvae+: Enhancing the generative quality of multimodal vaes without compromises}.
\newblock \emph{\bibinfo{journal}{Fifth Symposium on Advances in Approximate Bayesian Inference-Fast Track}}  (\bibinfo{year}{2023}).

\bibitem{sutter21B}
\bibinfo{author}{Sutter, T.~M.} \& \bibinfo{author}{Vogt, J.~E.}
\newblock \bibinfo{title}{Multimodal relational vae}.
\newblock \emph{\bibinfo{journal}{arXiv preprint}}  (\bibinfo{year}{2021}).

\bibitem{wolff21}
\bibinfo{author}{Wolff, J.} \emph{et~al.}
\newblock \bibinfo{title}{Hierarchical multimodal variational autoencoders}.
\newblock \emph{\bibinfo{journal}{arXiv preprint}}  (\bibinfo{year}{2021}).

\bibitem{vasco22}
\bibinfo{author}{Vasco, M.}, \bibinfo{author}{Yin, H.}, \bibinfo{author}{Melo, F.~S.} \& \bibinfo{author}{Paiva, A.}
\newblock \bibinfo{title}{Leveraging hierarchy in multimodal generative models for effective cross-modality inference}.
\newblock \emph{\bibinfo{journal}{Neural Networks}} \textbf{\bibinfo{volume}{146}}, \bibinfo{pages}{238--255} (\bibinfo{year}{2022}).

\bibitem{vaswani17}
\bibinfo{author}{Vaswani, A.} \emph{et~al.}
\newblock \bibinfo{title}{Attention is all you need}.
\newblock \emph{\bibinfo{journal}{Advances in Neural Information Processing Systems}} \textbf{\bibinfo{volume}{30}} (\bibinfo{year}{2017}).

\bibitem{ramesh22}
\bibinfo{author}{Ramesh, A.}, \bibinfo{author}{Dhariwal, P.}, \bibinfo{author}{Nichol, A.}, \bibinfo{author}{Chu, C.} \& \bibinfo{author}{Chen, M.}
\newblock \bibinfo{title}{Hierarchical text-conditional image generation with clip latents}.
\newblock \emph{\bibinfo{journal}{arXiv preprint arXiv:2204.06125}}  (\bibinfo{year}{2022}).

\bibitem{saharia22}
\bibinfo{author}{Saharia, C.} \emph{et~al.}
\newblock \bibinfo{title}{Photorealistic text-to-image diffusion models with deep language understanding}.
\newblock \emph{\bibinfo{journal}{Advances in Neural Information Processing Systems}} \textbf{\bibinfo{volume}{35}}, \bibinfo{pages}{36479--36494} (\bibinfo{year}{2022}).

\bibitem{radford21}
\bibinfo{author}{Radford, A.} \emph{et~al.}
\newblock \bibinfo{title}{Learning transferable visual models from natural language supervision}.
\newblock \emph{\bibinfo{journal}{International Conference on Machine Learning}} \bibinfo{pages}{8748--8763} (\bibinfo{year}{2021}).

\bibitem{bachmann22}
\bibinfo{author}{Bachmann, R.}, \bibinfo{author}{Mizrahi, D.}, \bibinfo{author}{Atanov, A.} \& \bibinfo{author}{Zamir, A.}
\newblock \bibinfo{title}{Multimae: Multi-modal multi-task masked autoencoders}.
\newblock \emph{\bibinfo{journal}{arXiv preprint arXiv:2204.01678}}  (\bibinfo{year}{2022}).

\bibitem{dosovitskiy20}
\bibinfo{author}{Dosovitskiy, A.} \emph{et~al.}
\newblock \bibinfo{title}{An image is worth 16x16 words: Transformers for image recognition at scale}.
\newblock \emph{\bibinfo{journal}{arXiv preprint arXiv:2010.11929}}  (\bibinfo{year}{2020}).

\bibitem{marino18}
\bibinfo{author}{Marino, J.}, \bibinfo{author}{Yue, Y.} \& \bibinfo{author}{Mandt, S.}
\newblock \bibinfo{title}{Iterative amortized inference}.
\newblock \emph{\bibinfo{journal}{International Conference on Machine Learning}} \bibinfo{pages}{3403--3412} (\bibinfo{year}{2018}).

\bibitem{kingma14}
\bibinfo{author}{Kingma, D.~P.} \& \bibinfo{author}{Ba, J.}
\newblock \bibinfo{title}{Adam: A method for stochastic optimization}.
\newblock \emph{\bibinfo{journal}{arXiv preprint arXiv:1412.6980}}  (\bibinfo{year}{2014}).

\bibitem{heusel2017gans}
\bibinfo{author}{Heusel, M.}, \bibinfo{author}{Ramsauer, H.}, \bibinfo{author}{Unterthiner, T.}, \bibinfo{author}{Nessler, B.} \& \bibinfo{author}{Hochreiter, S.}
\newblock \bibinfo{title}{Gans trained by a two time-scale update rule converge to a local nash equilibrium}.
\newblock \emph{\bibinfo{journal}{Advances in Neural Information Processing Systems}} \textbf{\bibinfo{volume}{30}} (\bibinfo{year}{2017}).

\bibitem{clevert2015elu}
\bibinfo{author}{Clevert, D.-A.}, \bibinfo{author}{Unterthiner, T.} \& \bibinfo{author}{Hochreiter, S.}
\newblock \bibinfo{title}{Fast and accurate deep network learning by exponential linear units (elus)}.
\newblock \emph{\bibinfo{journal}{arXiv preprint arXiv:1511.07289}}  (\bibinfo{year}{2015}).

\bibitem{ba2016layernorm}
\bibinfo{author}{Ba, J.~L.}, \bibinfo{author}{Kiros, J.~R.} \& \bibinfo{author}{Hinton, G.~E.}
\newblock \bibinfo{title}{Layer normalization}.
\newblock \emph{\bibinfo{journal}{arXiv preprint arXiv:1607.06450}}  (\bibinfo{year}{2016}).

\end{thebibliography}

\appendix

\section{Implementation Details of Multimodal Iterative Amortized Inference}
\label{implementation_details}
In our experiments, we implemented multimodal iterative inference models following the settings of~\citep{marino18}. 
The model iteratively updates the mean \(\boldsymbol{\mu}_m\) and log-variance \(\log \boldsymbol{\sigma}_m\) of the latent distribution by incorporating information from the input data \(\mathbf{x}_m\) and the gradients of the multimodal ELBO with respect to these parameters. 

At each iteration \(t\), the mean \(\boldsymbol{\mu}_{m_t}\) and log-variance \(\log \boldsymbol{\sigma}_{m_t}\) are updated based on the input data \(\mathbf{x}_m\), as well as the gradients \(\nabla_{\boldsymbol{\mu}_{m_t}} \mathcal{L}\) and \(\nabla_{\log \boldsymbol{\sigma}_{m_t}} \mathcal{L}\), using the following function:

\begin{equation}
\label{iai_applied}
\boldsymbol{\mu}_{m_{t+1}}, \boldsymbol{\sigma}_{m_{t+1}} = f_{w}\left(\mathbf{x}_m, \boldsymbol{\mu}_{m_t}, \log \boldsymbol{\sigma}_{m_t}, \nabla_{\boldsymbol{\mu}_{m_t}} \mathcal{L}, \nabla_{\log \boldsymbol{\sigma}_{m_t}} \mathcal{L} \right),
\end{equation}
where \(f_{w}\) represents the iterative amortized inference model.

Initially, we extract features from the input data \(\mathbf{x}_m\) by applying a linear transformation followed by the exponential linear unit (ELU) activation function~\citep{clevert2015elu}:

\[
\mathbf{h}_{\mathbf{x}} = \mathrm{ELU}\left( \mathbf{W}_{\mathbf{x}} \mathbf{x}_m + \mathbf{b}_{\mathbf{x}} \right),
\]
where \(\mathbf{W}_{\mathbf{x}}\) and \(\mathbf{b}_{\mathbf{x}}\) are learnable parameters. 

Next, the gradients of the multimodal ELBO with respect to the current mean and log-variance, \(\nabla_{\boldsymbol{\mu}_{m_t}} \mathcal{L}\) and \(\nabla_{\log \boldsymbol{\sigma}_{m_t}} \mathcal{L}\), are processed using layer normalization~\citep{ba2016layernorm} to stabilize the training:

\[
\tilde{\nabla}_{\boldsymbol{\mu}_{m_t}} \mathcal{L} = \mathrm{LayerNorm}\left( \nabla_{\boldsymbol{\mu}_{m_t}} \mathcal{L} \right), \quad
\tilde{\nabla}_{\log \boldsymbol{\sigma}_{m_t}} \mathcal{L} = \mathrm{LayerNorm}\left( \nabla_{\log \boldsymbol{\sigma}_{m_t}} \mathcal{L} \right).
\]

These normalized gradients are concatenated with the current estimates of \(\boldsymbol{\mu}_{m_t}\) and \(\log \boldsymbol{\sigma}_{m_t}\) to form a combined feature representation:

\[
\mathbf{h}_{\text{grad}} = \mathrm{ELU}\left( \mathbf{W}_{\text{grad}} \left[ \boldsymbol{\mu}_{m_t}, \log \boldsymbol{\sigma}_{m_t}, \tilde{\nabla}_{\boldsymbol{\mu}_{m_t}} \mathcal{L}, \tilde{\nabla}_{\log \boldsymbol{\sigma}_{m_t}} \mathcal{L} \right] + \mathbf{b}_{\text{grad}} \right),
\]
where \([ \cdot ]\) denotes concatenation, and \(\mathbf{W}_{\text{grad}}\) and \(\mathbf{b}_{\text{grad}}\) are learnable parameters.

The data feature \(\mathbf{h}_{\mathbf{x}}\) and the gradient feature \(\mathbf{h}_{\text{grad}}\) are concatenated to form a combined representation:

\[
\mathbf{h} = \left[ \mathbf{h}_{\mathbf{x}}, \mathbf{h}_{\text{grad}} \right].
\]

This combined feature vector is passed through additional neural network layers to compute candidate updates for the mean and log-variance:

\[
\tilde{\boldsymbol{\mu}}_m = \tanh\left( \mathbf{W}_{\mu} \mathbf{h} + \mathbf{b}_{\mu} \right), \quad
\tilde{\log \boldsymbol{\sigma}}_m = \tanh\left( \mathbf{W}_{\sigma} \mathbf{h} + \mathbf{b}_{\sigma} \right),
\]
where \(\tanh\) ensures bounded outputs, preventing numerical instability.

To determine the influence of these candidate updates on the current estimates, we apply gating functions computed with a sigmoid activation:

\[
\mathbf{g}_{\mu} = \mathrm{sigmoid}\left( \mathbf{W}_{\mu}^{\text{gate}} \mathbf{h} + \mathbf{b}_{\mu}^{\text{gate}} \right), \quad
\mathbf{g}_{\sigma} = \mathrm{sigmoid}\left( \mathbf{W}_{\sigma}^{\text{gate}} \mathbf{h} + \mathbf{b}_{\sigma}^{\text{gate}} \right).
\]

The updated mean and log-variance are computed as a combination of the current estimates and candidate updates, weighted by the gating mechanisms:

\[
\boldsymbol{\mu}_{m_{t+1}} = \mathbf{g}_{\mu} \odot \boldsymbol{\mu}_{m_t} + \left( 1 - \mathbf{g}_{\mu} \right) \odot \tilde{\boldsymbol{\mu}}_m,
\]

\[
\log \boldsymbol{\sigma}_{m_{t+1}} = \mathbf{g}_{\sigma} \odot \log \boldsymbol{\sigma}_{m_t} + \left( 1 - \mathbf{g}_{\sigma} \right) \odot \tilde{\log \boldsymbol{\sigma}}_m,
\]
where \(\odot\) represents element-wise multiplication.

By repeating this step for \(T\) iterations, we obtain the final updated values \(\boldsymbol{\mu}_{m_T}\) and \(\log \boldsymbol{\sigma}_{m_T}\), which define the approximate posterior distribution:

\[
q(\mathbf{z}_T | \mathbf{x}_m) = \mathcal{N}\left( \mathbf{z}_T; \boldsymbol{\mu}_{m_T}, \exp\left( \log \boldsymbol{\sigma}_{m_T} \right)^2 \right).
\]

This iterative process effectively incorporates multimodal information into the latent representation inferred from the unimodal input \(\mathbf{x}_m\), addressing both the amortization gap and the information loss caused by missing modalities.

\section{Visualization of Privious Models}
\begin{figure*}[t]
\centering
\subfloat[Mixture-based models]{\includegraphics[clip, width=0.9\linewidth]{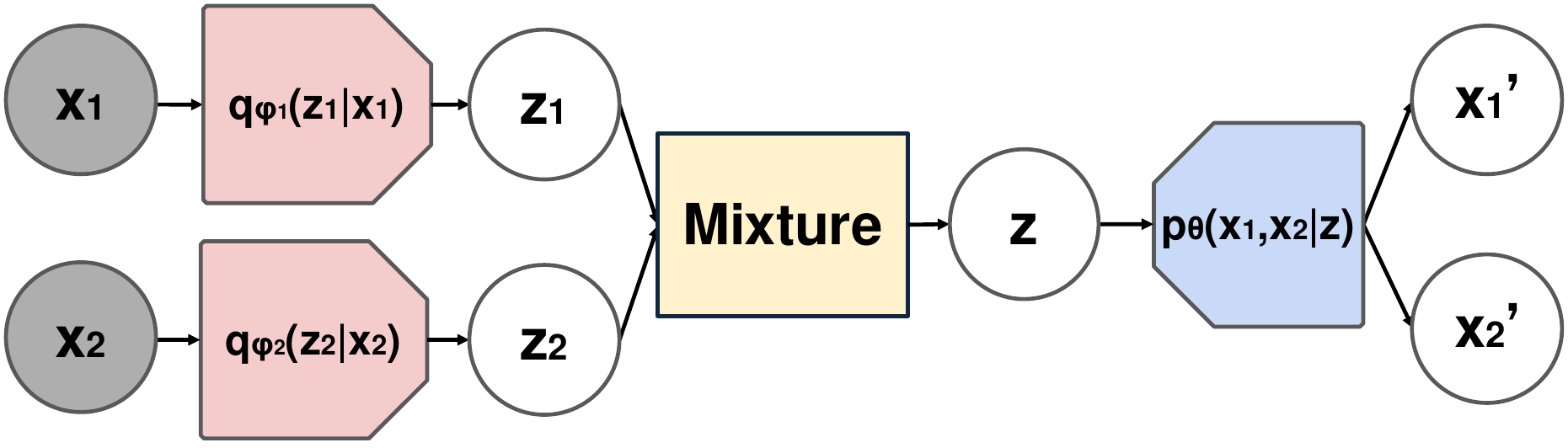}
\label{fig:mixture-based}}
\\
\subfloat[Alignment-based models]{\includegraphics[clip, width=\linewidth]{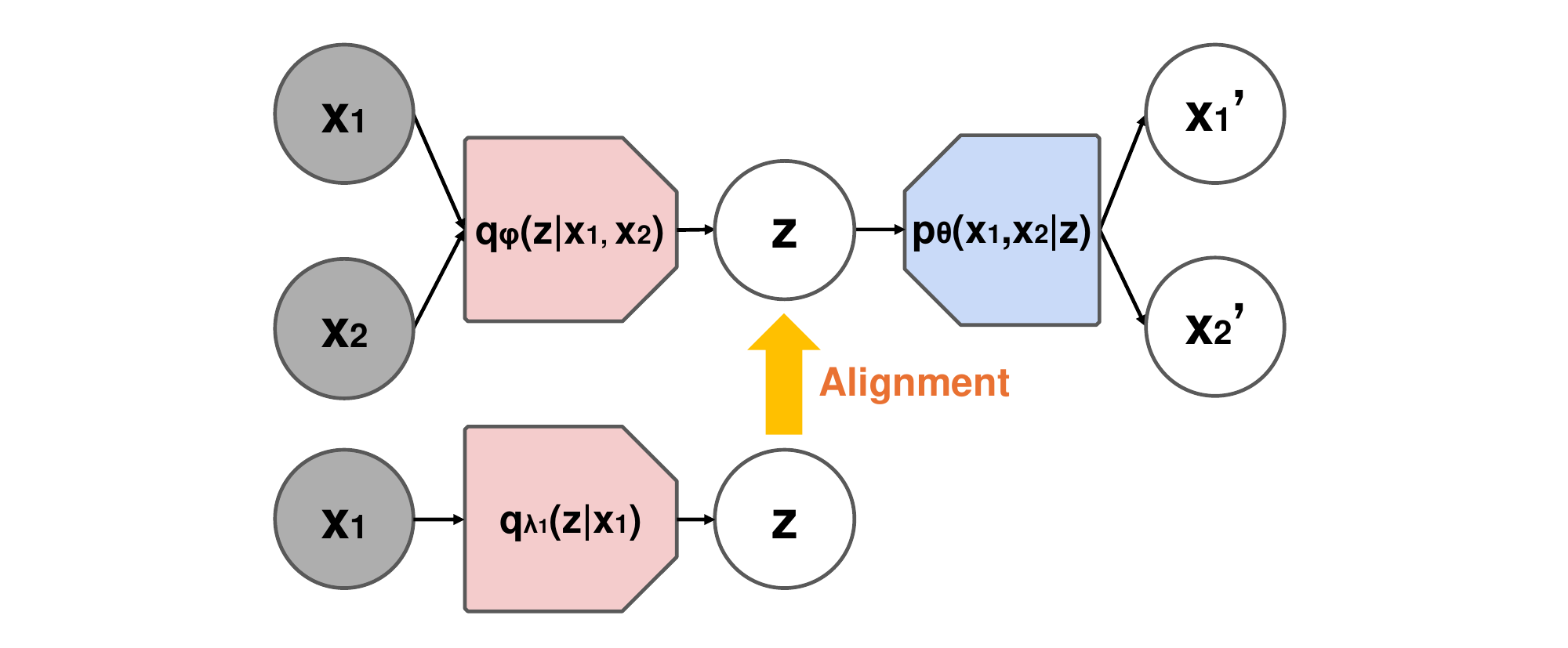}
\label{fig:alignment-based}}

\caption{Visualizatuon of previous multimodal VAEs (\# of modality types $M = 2$).}
\label{fig:previous}
\end{figure*}

In the method part, we graphically visualize proposed methods in Figure~\ref{fig:previous}.

\section{Additional Results}
In this Appendix, we provide additional results of our experiments.

\subsection{Further Qualitative Results of Multimodal Iterative Amortized Inference}
\label{teacher-qualitative-appendix}
\begin{figure}[tb]
\centering
\begin{minipage}[b]{0.49\columnwidth}
    \centering
    \includegraphics[width=1.0\columnwidth]{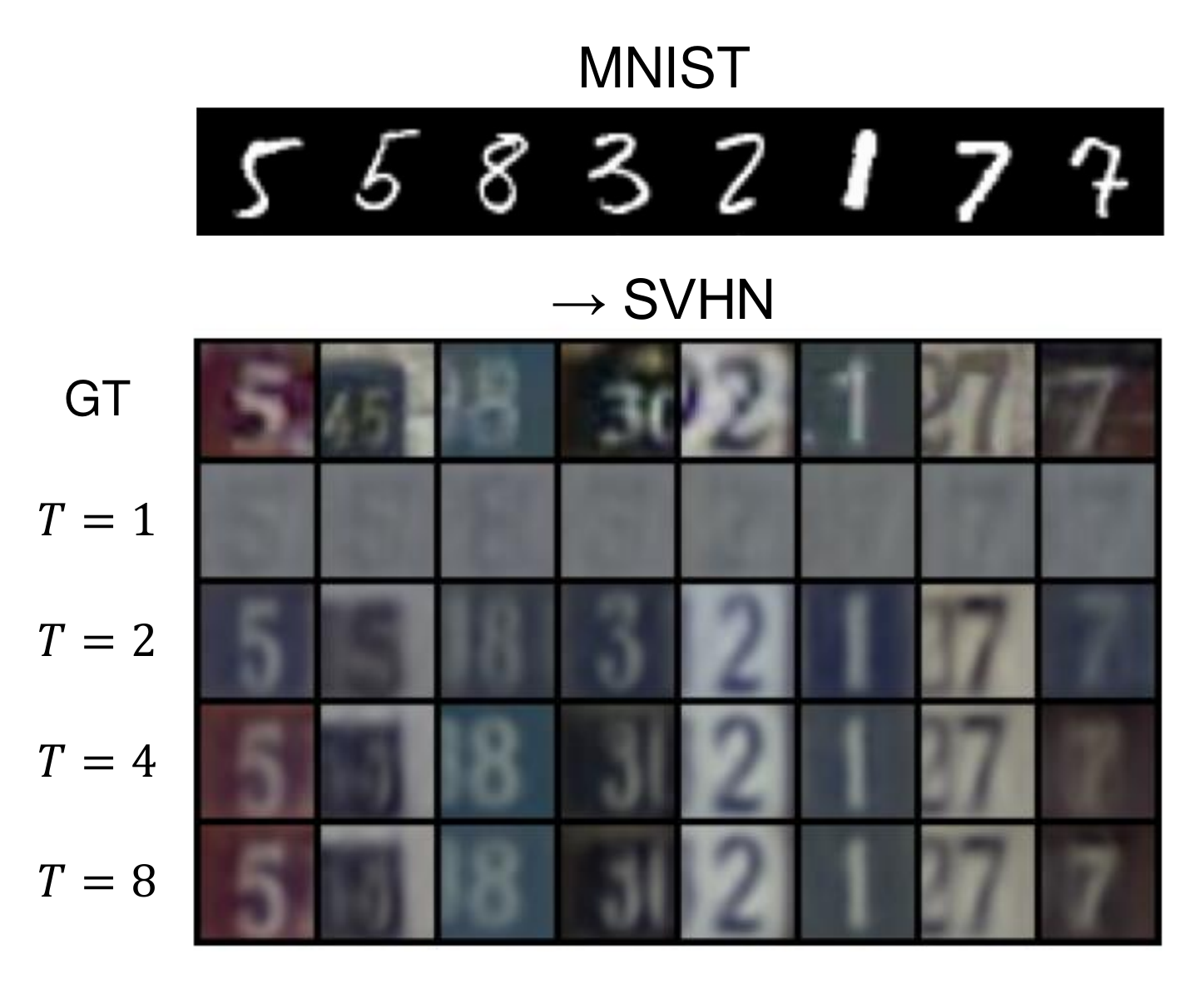}
\end{minipage}
\begin{minipage}[b]{0.49\columnwidth}
    \centering
    \includegraphics[width=1.0\columnwidth]{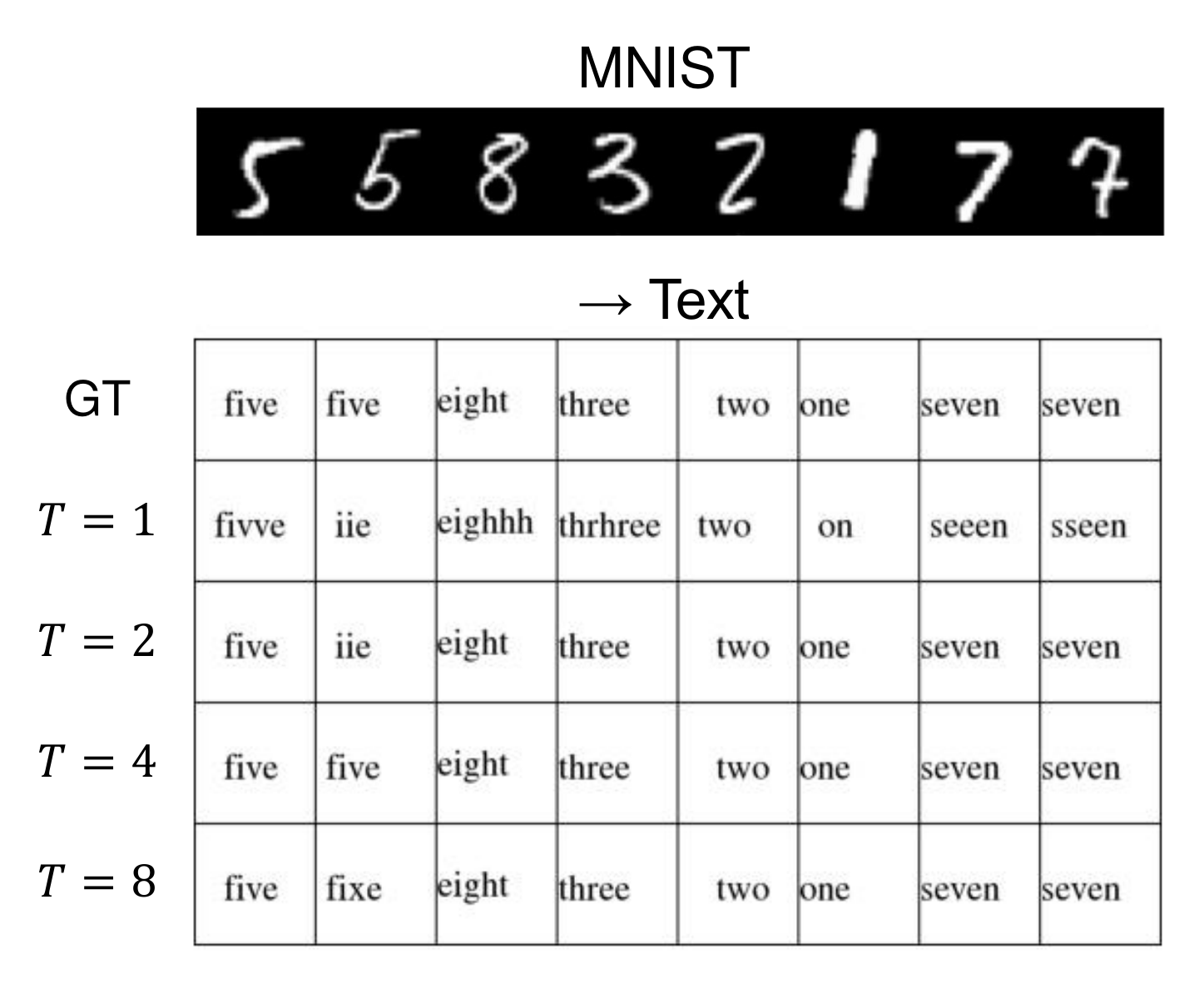}
\end{minipage}
\caption{Qualitative results of cross-modal generation on the MNIST-SVHN-Text dataset when applying multimodal iterative amortized inference to \( q_{\phi_0}(\mathbf{z}|\mathbf{x}_0) \) (input modality $\mathbf{x}_0$ is MNIST, generated modality $\mathbf{x}_1$ is SVHN (left), and generated modality $\mathbf{x}_2$ is Text (right)). By increasing the number of iterations \( T \), information from missing modalities is recovered, improving the performance of cross-modal generation.}
\label{iterative_generation_mst_0}  
\end{figure}

\begin{figure}[tb]
\centering
\begin{minipage}[b]{0.49\columnwidth}
    \centering
    \includegraphics[width=1.0\columnwidth]{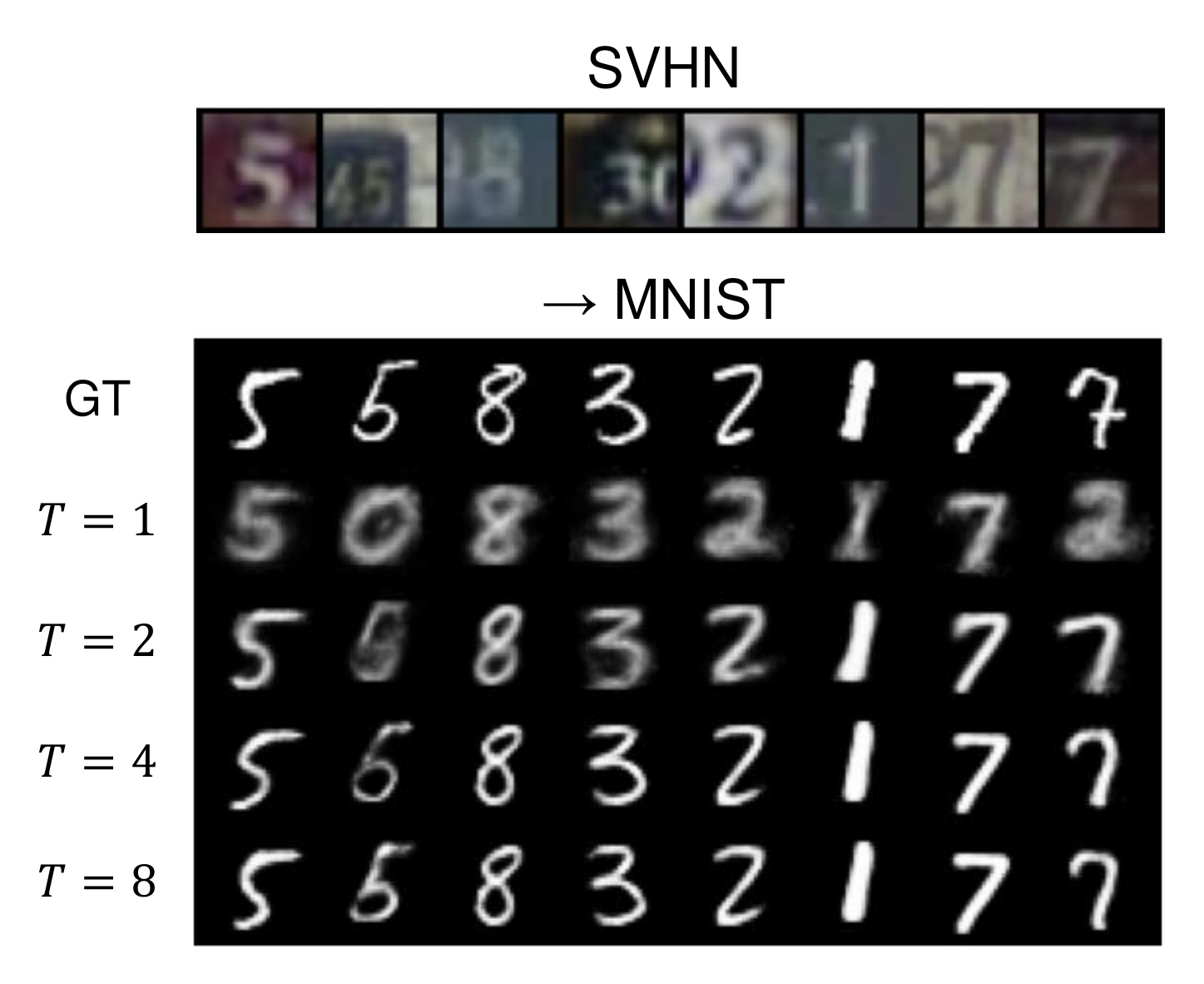}
\end{minipage}
\begin{minipage}[b]{0.49\columnwidth}
    \centering
    \includegraphics[width=1.0\columnwidth]{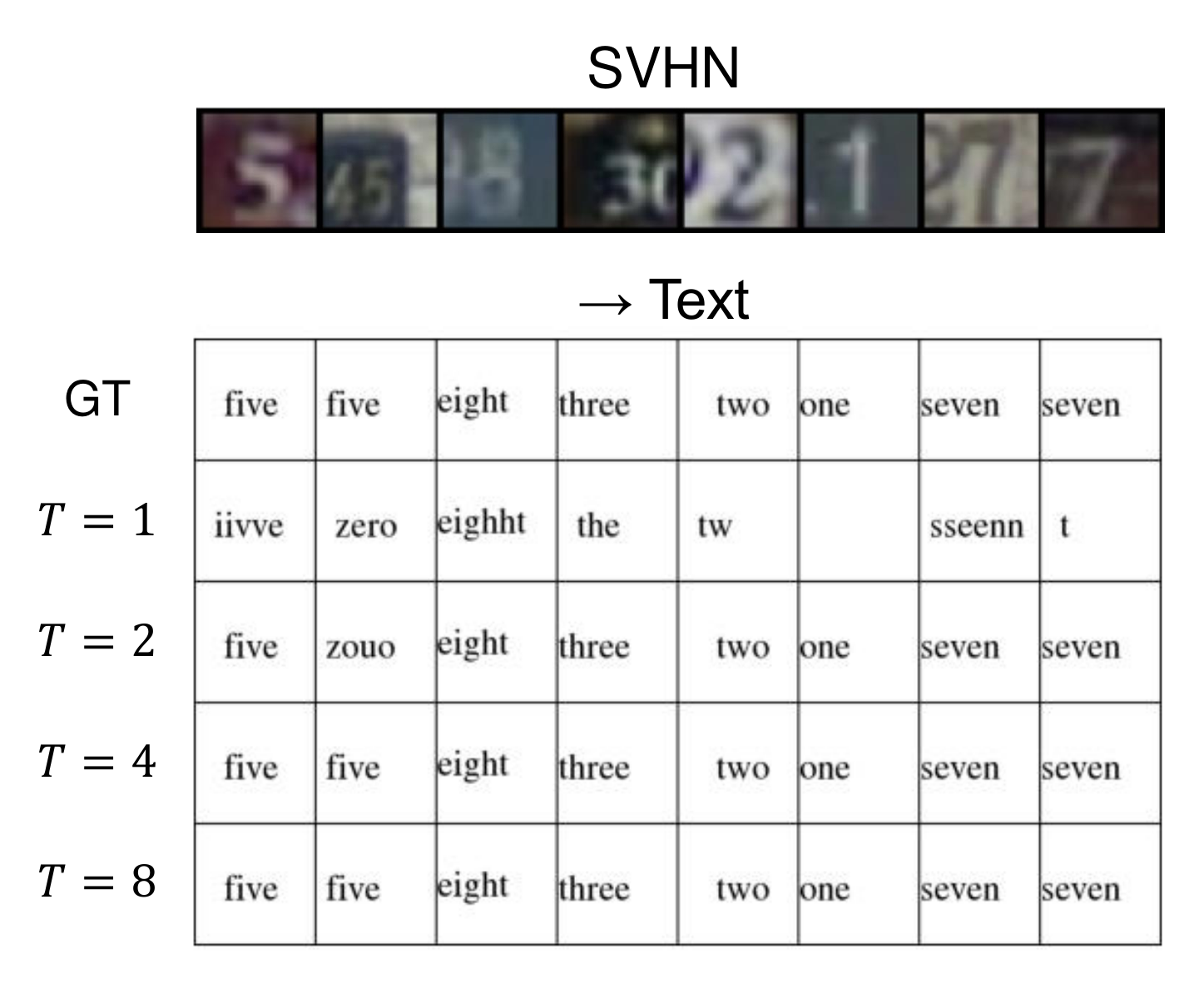}
\end{minipage}
\caption{Qualitative results of cross-modal generation on the MNIST-SVHN-Text dataset when applying multimodal iterative amortized inference to \( q_{\phi_1}(\mathbf{z}|\mathbf{x}_1) \) (input modality $\mathbf{x}_1$ is SVHN, generated modality $\mathbf{x}_0$ is MNIST (left), and generated modality $\mathbf{x}_2$ is Text (right)). By increasing the number of iterations \( T \), information from missing modalities is recovered, improving the performance of cross-modal generation.}
\label{iterative_generation_mst_1}  
\end{figure}

In the experiments and results part of the paper, we show qualitative performance of multimodal iterative amortized inference with MNIST-SVHN-Text dataset
in Figure~\ref{iterative_generation_mst_0}, Figure~\ref{iterative_generation_mst_1}. 
For the dataset used in this study, information from missing modalities is recovered by increasing the number of iterations $T$.

\subsection{Further Comparison with Previous Studies}
\label{student-qualitative-appendix}
\begin{figure}[tb]
\centering
\begin{minipage}[b]{0.49\columnwidth}
    \centering
    \includegraphics[width=1.0\columnwidth]{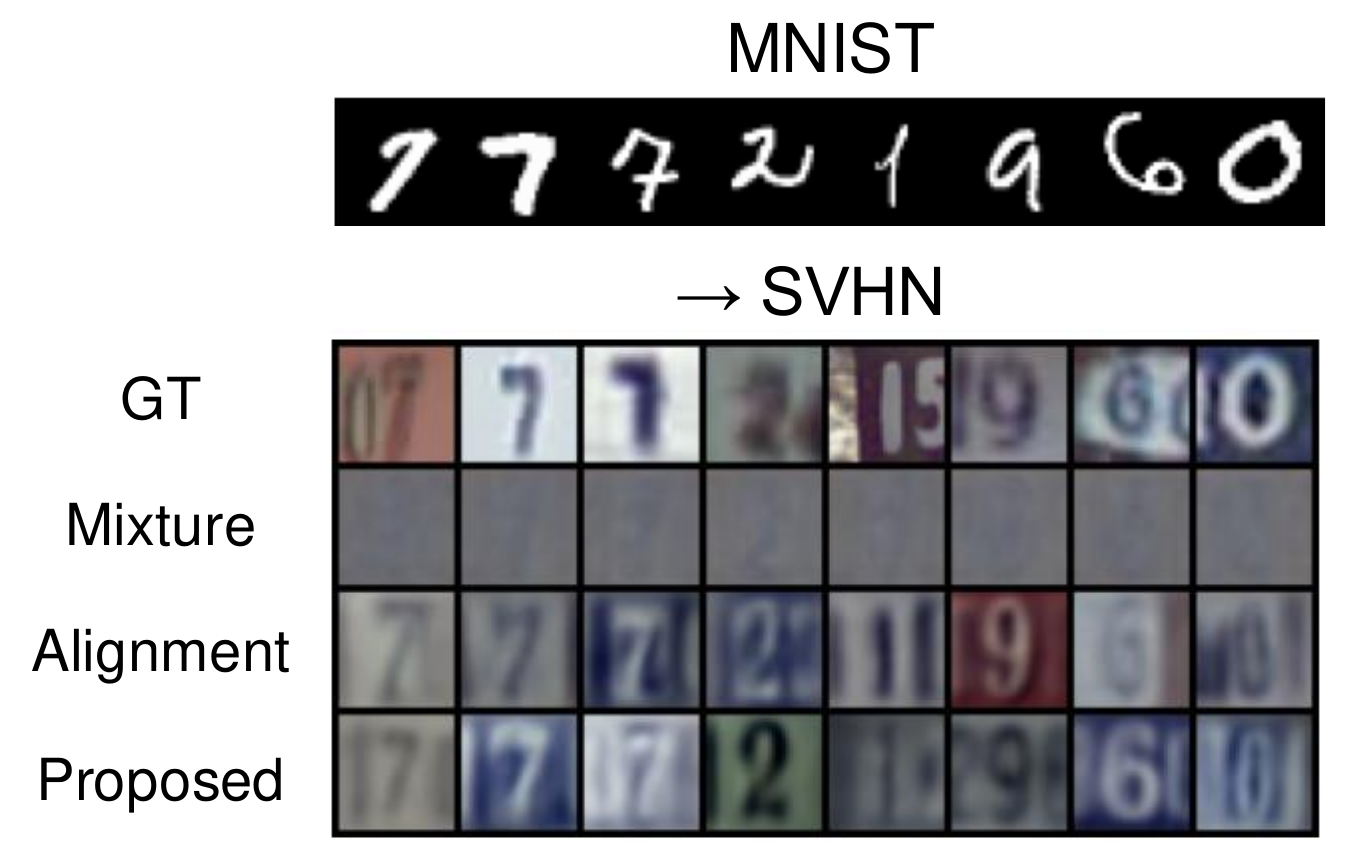}
\end{minipage}
\begin{minipage}[b]{0.49\columnwidth}
    \centering
    \includegraphics[width=1.0\columnwidth]{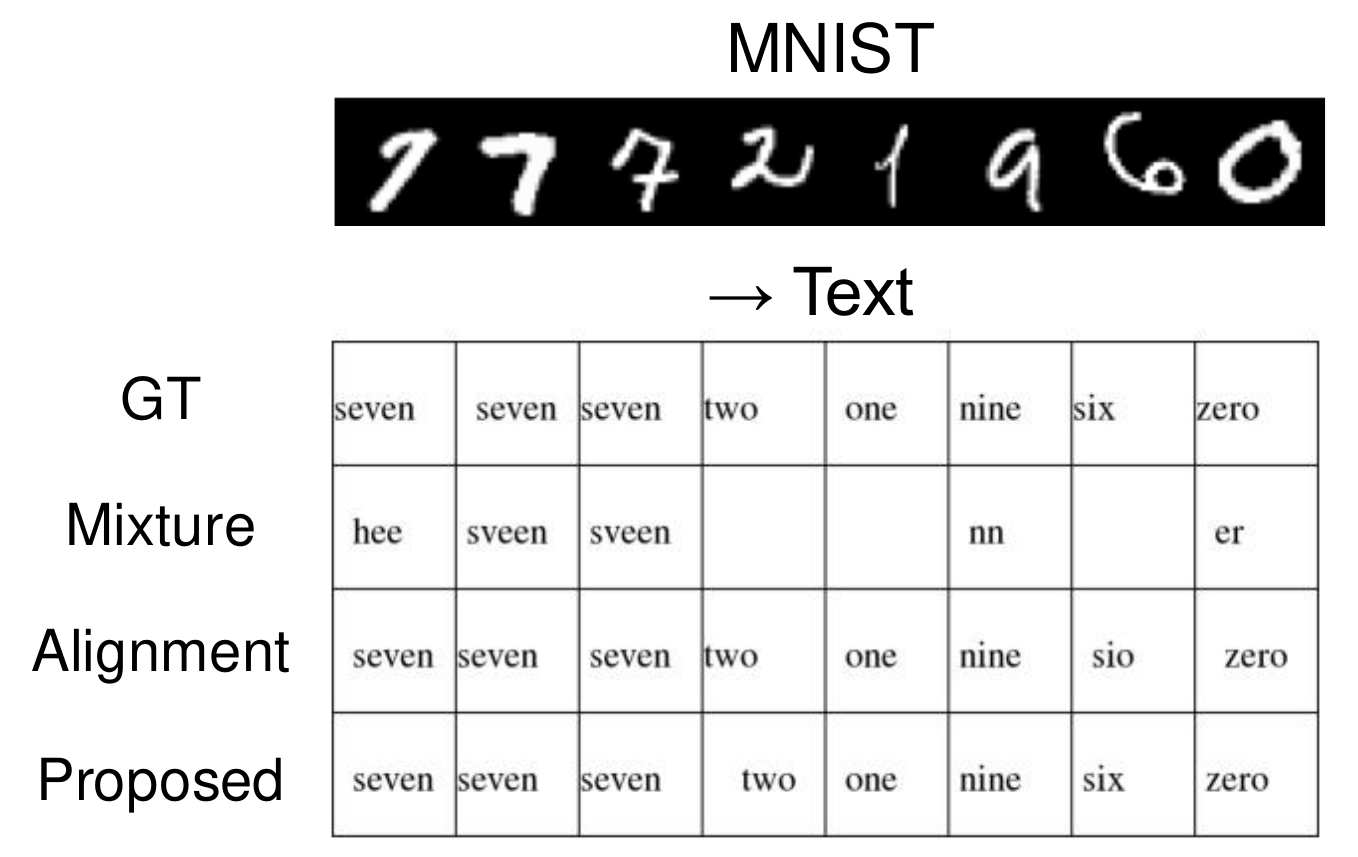}
\end{minipage}
\caption{Comparison of cross-modal generation on the MNIST-SVHN-Text dataset (input modality $\mathbf{x}_0$ is MNIST, generated modality $\mathbf{x}_1$ is SVHN (left), and generated modality $\mathbf{x}_2$ is Text (right)). }
\label{comparison_student_0}  
\end{figure}

\begin{figure}[tb]
\centering
\begin{minipage}[b]{0.49\columnwidth}
    \centering
    \includegraphics[width=1.0\columnwidth]{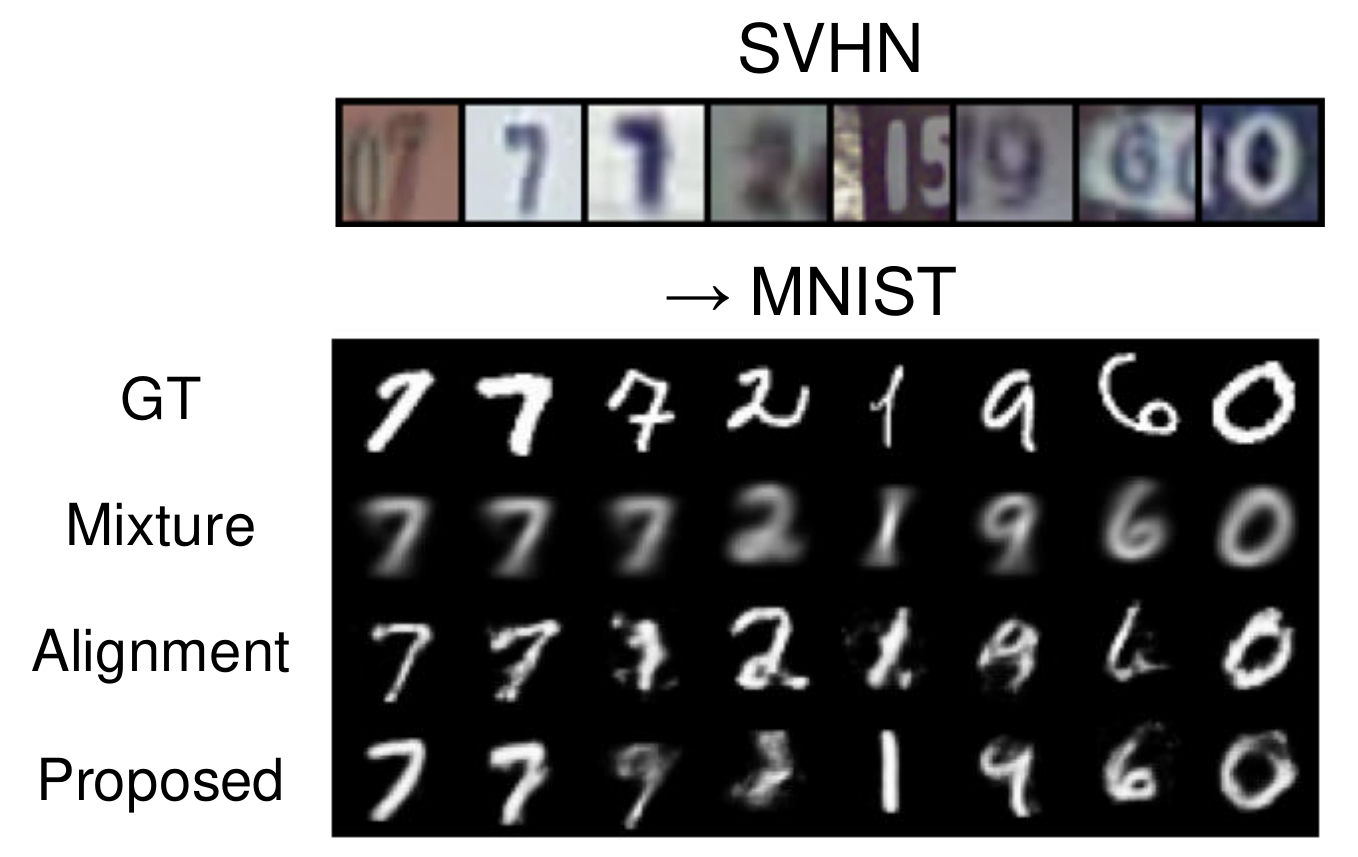}
\end{minipage}
\begin{minipage}[b]{0.49\columnwidth}
    \centering
    \includegraphics[width=1.0\columnwidth]{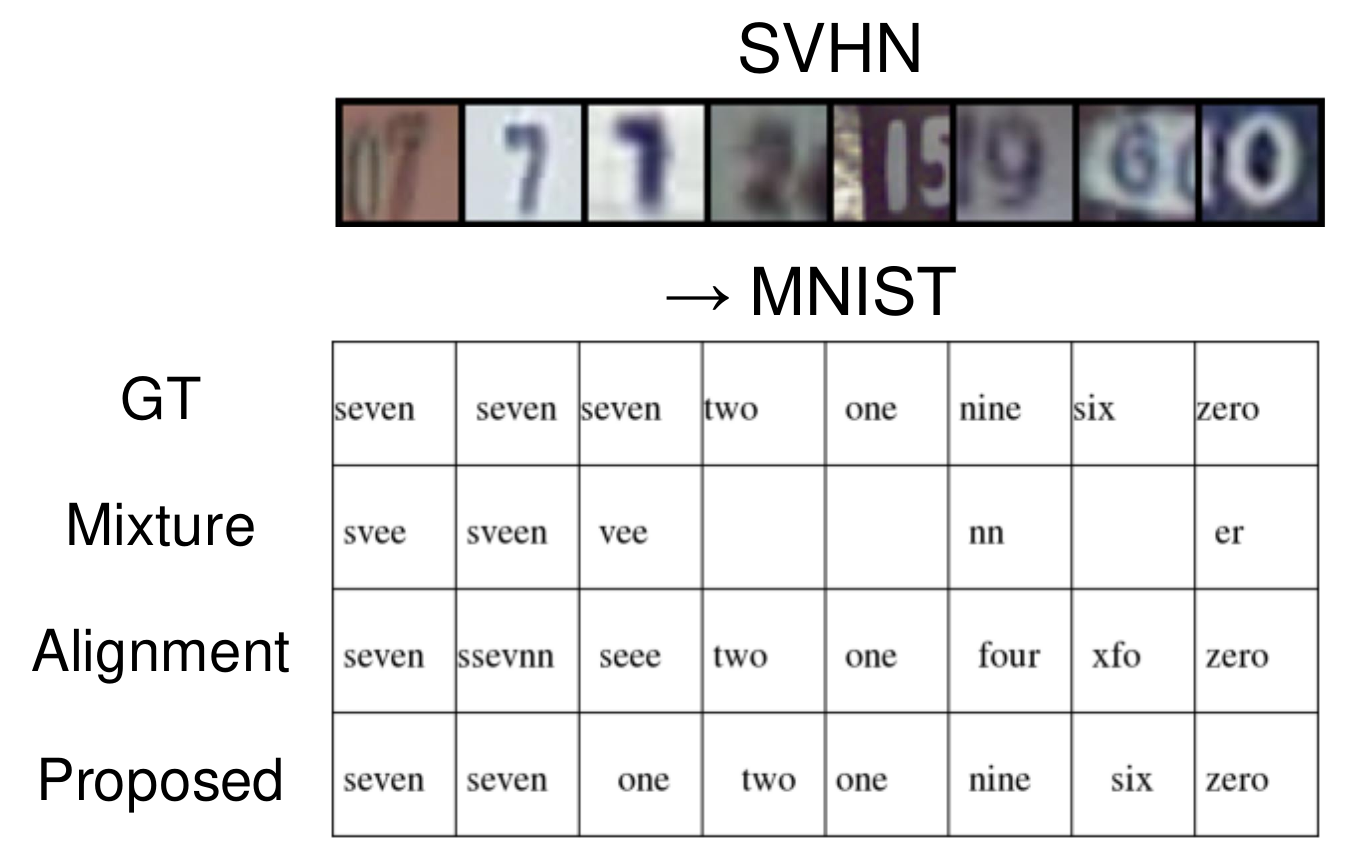}
\end{minipage}
\caption{Comparison of cross-modal generation on the MNIST-SVHN-Text dataset (input modality $\mathbf{x}_1$ is SVHN, generated modality $\mathbf{x}_0$ is MNIST (left), and generated modality $\mathbf{x}_2$ is Text (right)). }
\label{comparison_student_1}  
\end{figure}

In the experiments and results part of the paper, we show comparison with previosu studies in MNIST-SVHN-Text dataset
in Figure~\ref{comparison_student_0}, Figure~\ref{comparison_student_1}. 
For the dataset used in this study, information from missing modalities is recovered by increasing the number of iterations $T$.

\end{document}